\documentclass[runningheads]{llncs}

\usepackage[final,year=2026,ID=4558]{eccv}

\usepackage{eccvabbrv}

\usepackage{graphicx}
\usepackage{booktabs}
\usepackage{amssymb}
\usepackage{pifont}
\usepackage{multirow}
\usepackage{multicol}
\usepackage{booktabs} 
\usepackage{array}    
\usepackage[table]{xcolor}
\usepackage{tabularx}
\usepackage{wrapfig}
\usepackage{microtype}
\usepackage{graphicx}
\usepackage[table]{xcolor} 
\usepackage{colortbl}
\definecolor{tableblue}{RGB}{235, 245, 255} 
\definecolor{tablegray}{gray}{0.95}         
\definecolor{tableblue}{RGB}{200, 220, 245}
\definecolor{tablepink}{RGB}{255, 200, 200}
\definecolor{tablegrey}{RGB}{230, 230, 230}
\newcommand{\cmark}{\ding{51}} 
\newcommand{\xmark}{\ding{55}} 
\usepackage[utf8]{inputenc}
\usepackage{booktabs, multirow, xcolor, colortbl, graphicx}
\usepackage{siunitx} 
\usepackage{xfp}     
\usepackage{amssymb} 
\newcommand{\hc}[1]{%
    \if\relax\detokenize{#1}\relax
        {#1}%
    \else
        \edef\grad{\fpeval{#1*40}}%
        \cellcolor{blue!\grad!white}#1%
    \fi
}
\newcommand{\gr}[2]{\if\relax\detokenize{#2}\relax -- \else \cellcolor{black!#1!white}#2 \fi}
\newcommand{\bl}[2]{\if\relax\detokenize{#2}\relax -- \else \cellcolor{blue!#1!white}#2 \fi}
\usepackage{xcolor}
\usepackage{colortbl} 

\newcommand{\pos}[1]{\textcolor{blue!70!black}{\scriptsize{(+#1)}}} 
\newcommand{\negv}[1]{\textcolor{red!70!black}{\scriptsize{(-#1)}}}  
\usepackage[accsupp]{axessibility}  
\usepackage{hyperref}

\usepackage{orcidlink}

\begin{document}
\pdfoutput=1
\title{Beyond Language: Grounding Referring Expressions with Hand Pointing in Egocentric Vision} 

\titlerunning{Abbreviated paper title}

\author{Ling Li\inst{1} \and
Bowen Liu\inst{2} \and
Zinuo Zhan\inst{3} \and
Peng Jie\inst{3} \and
Jianhui Zhong\inst{2} \and
Kenglun Chang\inst{4} \and
Zhidong Deng\inst{1,*}
}

\authorrunning{L. Li et al.}

\institute{Tsinghua University, Beijing, China \\
\email{liling25@mails.tsinghua.edu.cn} \and
Dalian University of Technology, Dalian, China \and
Northwestern Polytechnical University, Xi'an, China \and
Apple, USA \\
\vspace{2pt} 
\small{(*) Corresponding Author}
}

\maketitle

\begin{abstract}
  Traditional Visual Grounding (VG) predominantly relies on textual descriptions to localize objects, a paradigm that inherently struggles with linguistic ambiguity and often ignores non-verbal deictic cues prevalent in real-world interactions. In natural egocentric engagements, hand-pointing combined with speech forms the most intuitive referring mechanism. To bridge this gap, we introduce EgoPoint-Ground, the first large-scale multimodal dataset dedicated to egocentric deictic visual grounding. Comprising over \textbf{15k} interactive samples in complex scenes, the dataset provides rich, multi-grained annotations including hand-target bounding box pairs and dense semantic captions. We establish a comprehensive benchmark for hand-pointing referring expression resolution, evaluating a wide spectrum of mainstream Multimodal Large Language Models (MLLMs) and state-of-the-art VG architectures. Furthermore, we propose SV-CoT, a novel baseline framework that reformulates grounding as a structured inference process, synergizing gestural and linguistic cues through a Visual Chain-of-Thought paradigm. Extensive experiments demonstrate that SV-CoT achieves an $\textbf{11.7\%}$ absolute improvement over existing methods, effectively mitigating semantic ambiguity and advancing the capability of agents to comprehend multimodal physical intents. The dataset and code will be made publicly available.
\end{abstract}

\section{Introduction}
\label{sec:intro}
\begin{figure*}[t]
    \centering
    \includegraphics[width=\textwidth]{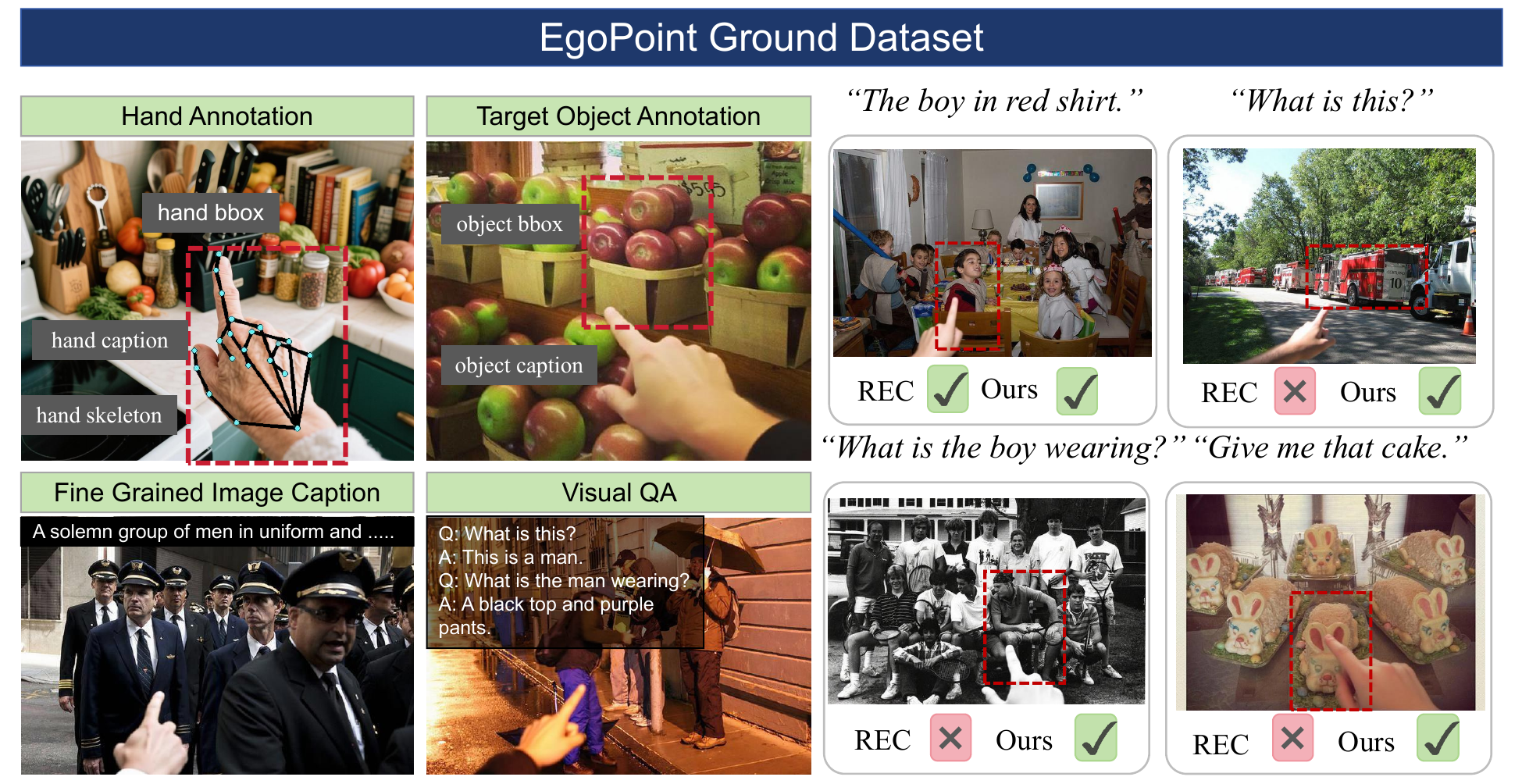}
    \caption{Our dataset features images captured from an egocentric perspective where a hand is pointing towards a target object. The dataset includes four core annotations: Hand Annotation, Target Object Annotation, Image Caption, and Visual QA Pairs. While conventional Referring Expression Comprehension (REC) tasks only support language-assisted localization, our proposed task aims to address referring disambiguation by utilizing physical hand information as opposed to traditional linguistic assistance for target grounding.}
    \label{fig:dataset egopoint}
\end{figure*}
Visual Grounding (VG) \cite{zhang2018grounding,dai2024simvg,xiao2024oneref,zhuang2018parallel,xiao2024towardssurvey,anderson2018vision,visualgrounding2_chen2024efficient,visualgrounding5_mo2024nexusad} aims to localize target objects in images based on natural language queries\cite{aim1_xiao2025towards,aim2_kang2025your,aim3_liu2025aerialvg}, serving as a fundamental capability for natural human-AI interaction\cite{pfeifer2004embodied,liu2025aligningembodied,duan2022surveyembodied,wang2023voyager,humanai1_jiang2024visual,humanai2_qi2024shapellm,humanai3_yang2024v}. Existing VG methods primarily rely on static third-person perspective images and purely textual descriptions for object localization~\cite{zhang2018grounding,liu2024groundingdino,you2023ferret,chen2023shikra,cheng2024yoloworld,nagaraja2016refcocog,wang2025internvl3,thirdperson1_huang2024loa,thirdperson3_kang2024segvg}. However, this paradigm faces fundamental challenges in real-world interactive scenarios: language inherently exhibits ambiguity\cite{ambiguity1_kusnanti2024indonesian,ambiguity2_fortuny2024ambiguity}, while humans heavily depend on non-linguistic cues (e.g., gestures) for referential disambiguation in daily communication\cite{bansal2022egoprocess,grauman2022Ego4d,grauman2024Ego-exo4d,efron1969perception,gesture1&mechanism1&diff1_shi2022spatial,gesture3_delmas2024poseembroider,gesture4_ju2024robo}. For instance, when a user says "pass me that one" models struggle to identify the intended referent without gestural context. Figure \ref{fig:dataset egopoint} illustrates the significant difference between our proposed method and current approaches on the referring question-answering task\cite{gesture1&mechanism1&diff1_shi2022spatial,diff2&applicability2_xie2024target}.

In physical interactions, the synergy between hand gestures and language constitutes the most natural and efficient referring mechanism\cite{chandel2015occlusion,hashi2024systematichand,gesture1&mechanism1&diff1_shi2022spatial,mechanism5_memmi2024hand}. A typical scenario involves a user wearing AR glasses\cite{arglasses1&guide1_lee2024gazepointar,arglasses2_bhattacharyya2024helios,arglasses3_magay2024light} pointing at an object and asking "What is this?"—the finger direction directly resolves linguistic ambiguity\cite{resolve1_jirak2021solving,resolve2&ar3_constantin2022interactive} and precisely guides attention\cite{arglasses1&guide1_lee2024gazepointar,guide2_watanabe2002reflexive} to the target\cite{2010referAt,guo2021humanhand,hashi2024systematichand,resolve2&ar3_constantin2022interactive}. Despite the ubiquity of such multimodal referring in human interaction, existing VG research has largely overlooked gestural cues in first-person perspectives\cite{chen2021yourefit,overlook2_ahuja2022communication}. Mainstream datasets focus on third-person images with descriptive text, lacking modeling of gesture-language coordination in egocentric interactions\cite{liu2024groundingdino,li2024groundinggpt,cheng2024yoloworld,chen2021yourefit,2010referAt,schauerte2010pointAT,lackego1_deichler2024mm,lackego2_farkhondeh2024childplay,lackego3_hummel2024egocvr}. Accordingly, existing benchmark tasks have not incorporated gestures into evaluation protocols, limiting model applicability in embodied AI\cite{applicability1_fan2024benchmarks,diff2&applicability2_xie2024target} and human-robot collaboration scenarios\cite{collaboration1_foteinos2026fr,collaboration2_hong2024enhancing}.

To bridge this gap, we introduce EgoPoint-Ground—the first large-scale multimodal dataset for first-person hand-pointing referring tasks. The dataset comprises over 15,000 high-quality samples captured from natural interactive moments in complex scenes. Each sample is annotated with hand bounding boxes, target object bounding boxes, image captions, object captions, category labels, and referring question-answer pairs, providing fine-grained supervision signals for multimodal referring expression resolution. Based on this dataset, we establish a benchmark for hand-pointing referring expression resolution and conduct systematic evaluations of mainstream Multimodal Large Language Models (MLLMs)\cite{bai2023qwen,yang2025qwen3,wang2024qwen2-vl,liu2024deepseek,guo2025seed1,team2023gemini,achiam2023gpt,chuang2025meta,zhang2024llava,an2025llavaoneversion} and visual grounding models. Experimental results demonstrate that existing methods suffer significant performance degradation when gestural cues are absent, quantitatively validating the critical role of hand gestures in resolving linguistic ambiguity.

Furthermore, we propose a novel baseline framework, SV-CoT, which explicitly synergizes gestural and linguistic cues through a Visual Chain-of-Thought reasoning paradigm. The model iteratively reasons about the spatial relationship between finger direction and candidate objects, effectively mitigating ambiguity introduced by purely linguistic descriptions. Experiments on EgoPoint-Ground show that our method achieves a 11.7$\%$ performance improvement over language-only baselines, fully demonstrating the necessity and effectiveness of multimodal fusion in real-world interactive scenarios.

Our main contributions are summarized as follows: \textbf{(1) Dataset:} We present EgoPoint-Ground, the first large-scale multimodal dataset for first-person hand-pointing referring, containing 15,000+ samples with multi-dimensional annotations including hand poses, object locations, and referring expressions. \textbf{(2) Benchmark:} We establish an evaluation benchmark for hand-pointing referring expression resolution and systematically analyze the limitations of existing state-of-the-art models on this task. \textbf{(3) Model:} We design a partial-Reasoning-based Visual Chain-of-Thought (SV-COT) baseline model, which fuses gestural and linguistic information via a Visual Chain-of-Thought mechanism, significantly improving referring localization accuracy. \textbf{(4) Open-source:} We release the dataset, benchmark code, and model implementation to advance research in multimodal referring and embodied AI.
\begin{figure*}
    \centering
    \includegraphics[width=\textwidth]{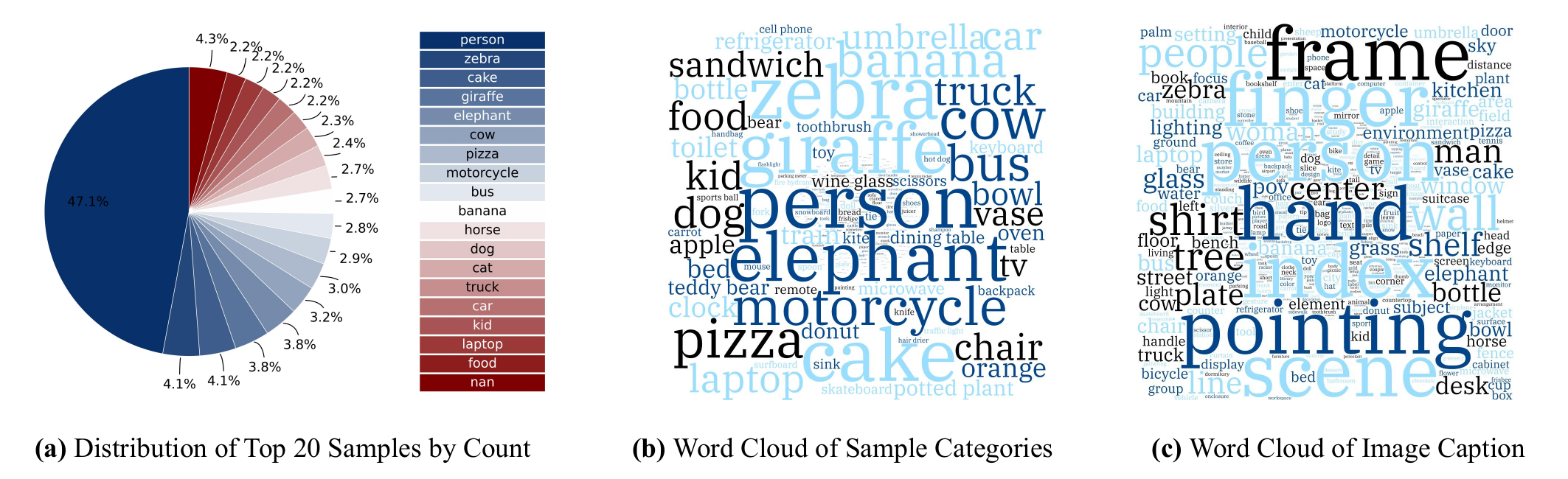}
    \caption{Statistical overview and distribution analysis of the Ego-Point Ground dataset. The dataset encompasses a broad spectrum of everyday objects, accompanied by semantically rich and lexically diverse textual annotations.}
    \label{fig:statistic of dataset-word cloud}
\end{figure*}


\section{EgoPoint Ground Dataset}
\label{sec:dataset}
\begin{table*}[t] 
    \centering
    \caption{Feature comparison of EgoPoint versus representative reference datasets.}
    \label{tab:dataset_comparison}
    
    \resizebox{\textwidth}{!}{
    \begin{tabular}{@{}lccccccccl@{}} 
        \toprule
        \textbf{Dataset} & \textbf{Lang.} & \textbf{Gest.} & \textbf{Embo.} & \textbf{Ego/Exo} & \textbf{Pose} & \textbf{VQA} & \textbf{Source} & \textbf{No. images}\\
        \midrule
        
        PointAt \cite{schauerte2010pointAT} & \xmark & \cmark & \cmark & Ego & \xmark & \xmark & Lab & 220 \\ 
        ReferAt \cite{2010referAt} & \xmark & \cmark & \cmark & Exo & \xmark & \xmark & Lab & 242 \\ 
        IPO \cite{shukla2015probabilisticIPO} & \xmark & \cmark & \cmark & Exo & \xmark & \xmark & Lab & 278 \\ 
        IMHF \cite{shukla2016multiIMHF} & \xmark & \xmark & \cmark & Ego & \xmark & \xmark & Lab & 1,716 \\ 
        RefIt \cite{kazemzadeh2014referitgamerefit} & \cmark & \xmark & \xmark & Exo & \xmark & \xmark & CLEF & 19,894 \\ 
        YourRefIt \cite{chen2021yourefit} & \cmark & \cmark & \cmark & Exo & \xmark & \xmark & Crowd-Sourced & 497,348 \\ 
        RefCOCO \cite{yu2016refcoco} & \cmark & \xmark & \xmark & Exo & \xmark & \xmark & MSCOCO & 19,994 \\ 
        RefCOCO+ \cite{mao2016refcoco+} & \cmark & \xmark & \xmark & Exo & \xmark & \xmark & MSCOCO & 19,992 \\ 
        RefCOCOg \cite{nagaraja2016refcocog} & \cmark & \xmark & \xmark & Exo & \xmark & \xmark & MSCOCO & 26,711 \\ 
        Flickr30k ent. \cite{plummer2015flickr30k} & \cmark & \xmark & \xmark & Exo & \xmark & \xmark & Flickr30K & 31,783 \\ 
        GuessWhat? \cite{de2017guesswhat} & \cmark & \xmark & \xmark & Exo & \xmark & \xmark & MSCOCO & 66,537 \\ 
        Cops-Ref \cite{chen2020cops-ref} & \xmark & \cmark & \xmark & Exo & \xmark & \xmark & COCO/Flickr & 75,299 \\ 
        CLEVR-Ref+ \cite{liu2019clevr-ref+} & \xmark & \xmark & \xmark & Exo & \xmark & \xmark & CLEVR & 99,992 \\ 
        \midrule
        
        \rowcolor{gray!10} 
        \textbf{Ours} & \cmark & \cmark & \cmark & \textbf{Ego} & \cmark & \cmark & \textbf{Mixed/Hybrid} & \textbf{15,338} \\
        
        \bottomrule
    \end{tabular}
    } 

    \vspace{4pt}
    \begin{minipage}{\textwidth}
        \scriptsize
        \textbf{Note:} Ego: Egocentric; Exo: Exocentric; Lang.: Language; Gest.: Gesture; Pose: Pose annotations; VQA: Visual Question Answering; Embo.: Embodied. \textit{Source} indicates the data origin: Lab (expert capture), public datasets (e.g., MSCOCO, CLEVR), or Crowd-sourced (AMT). Mixed/Hybrid denotes a combined collection methodology.
    \end{minipage}
\end{table*}
The EgoPoint-Ground Dataset represents the first high-fidelity and high-complexity egocentric benchmark designed explicitly for hand referential grounding. It features diverse scenarios critical for egocentric interaction, particularly simulating the challenge of referential Question Answering (QA) \cite{antol2015vqa} within smart-glasses settings \cite{waisberg2024metaglass}. The images are meticulously annotated with precise pixel-level joint labels, encompassing a comprehensive set of visual elements: target object bounding boxes, detailed hand poses information, and corresponding natural language descriptions. Table \ref{tab:dataset_comparison} details the comparison between our dataset and existing reference understanding datasets.

Figure \ref{fig:dataset egopoint} provides an overview of the dataset's structure and content. The following sections detail the composition, creation pipeline, and statistical analysis of the EgoPoint-Ground Dataset. More details are in the Appendix.

\subsection{Dataset Analysis}
\begin{figure*}
    \centering
    \includegraphics[width=0.9\textwidth]{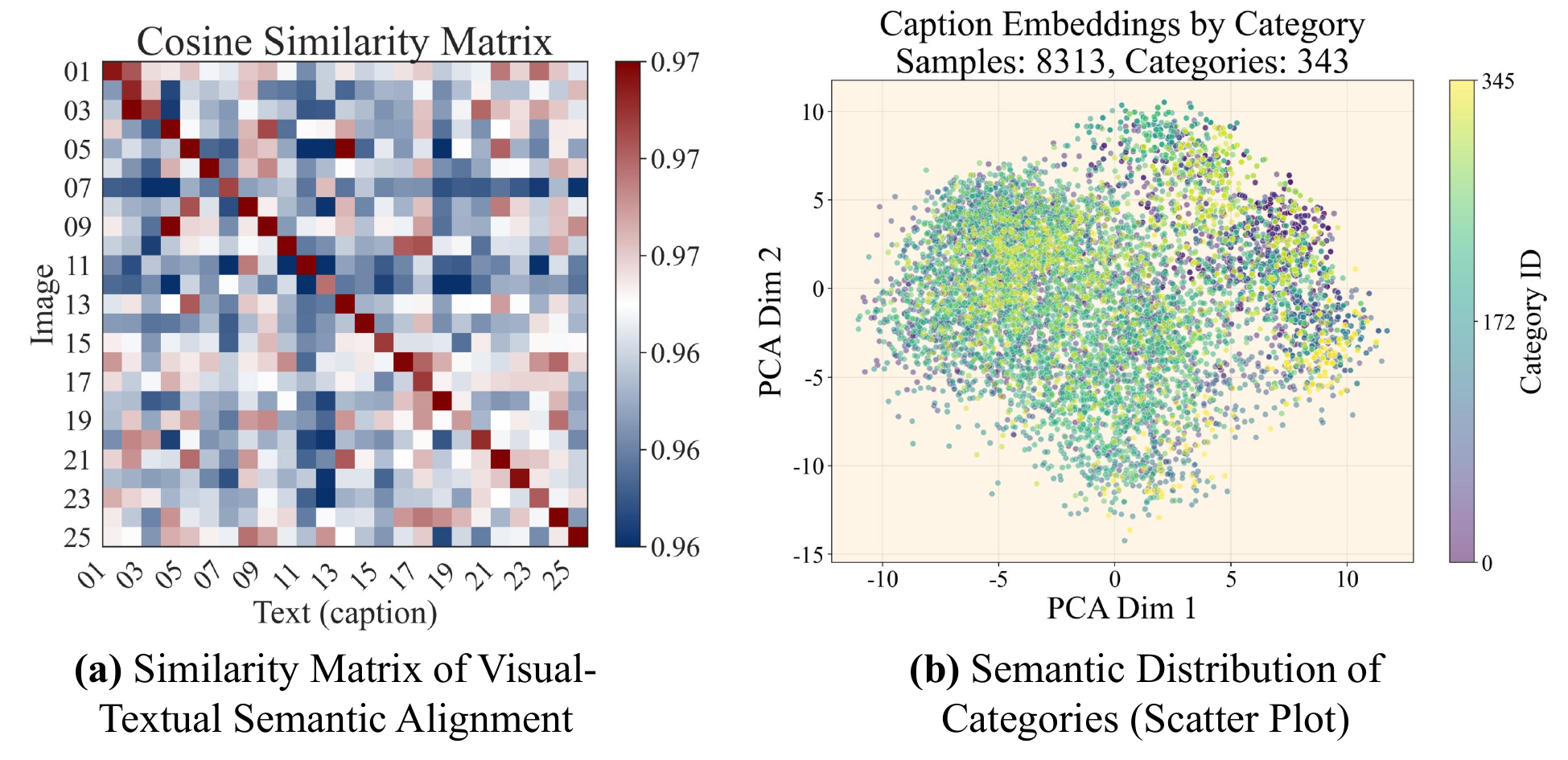}
    \caption{Analysis of Cross-Modal Quality and Semantic Diversity. Figure (a) shows the visual-textual semantic alignment heatmap. The high diagonal similarity confirms strong semantic consistency between images and descriptions. Figure (b) displays the category semantic distribution scatter plot. Image description embeddings are visualized in 2D using PCA (color-coded by category). The points are uniformly distributed without significant clustering, demonstrating the diversity of the dataset's semantic space.}
    \label{fig:annotation analyze-heatmap and point}
\end{figure*}
Our dataset comprises $15{,}338$ images spanning $140$ object categories and over $20$ everyday scenarios, including indoor, outdoor, animal, food, street, and kitchen environments. Each image contains an average of $2.8$ candidate objects, with $63.7\%$ of samples featuring $\geq 2$ objects of the same category and $42.5\%$ of targets exhibiting moderate or severe occlusion. The dataset includes both positive and negative samples, with $2{,}171$ negative samples ($14.15\%$) that further increase localization difficulty.
\begin{figure*}[t]
    \centering
    \includegraphics[width=\textwidth]{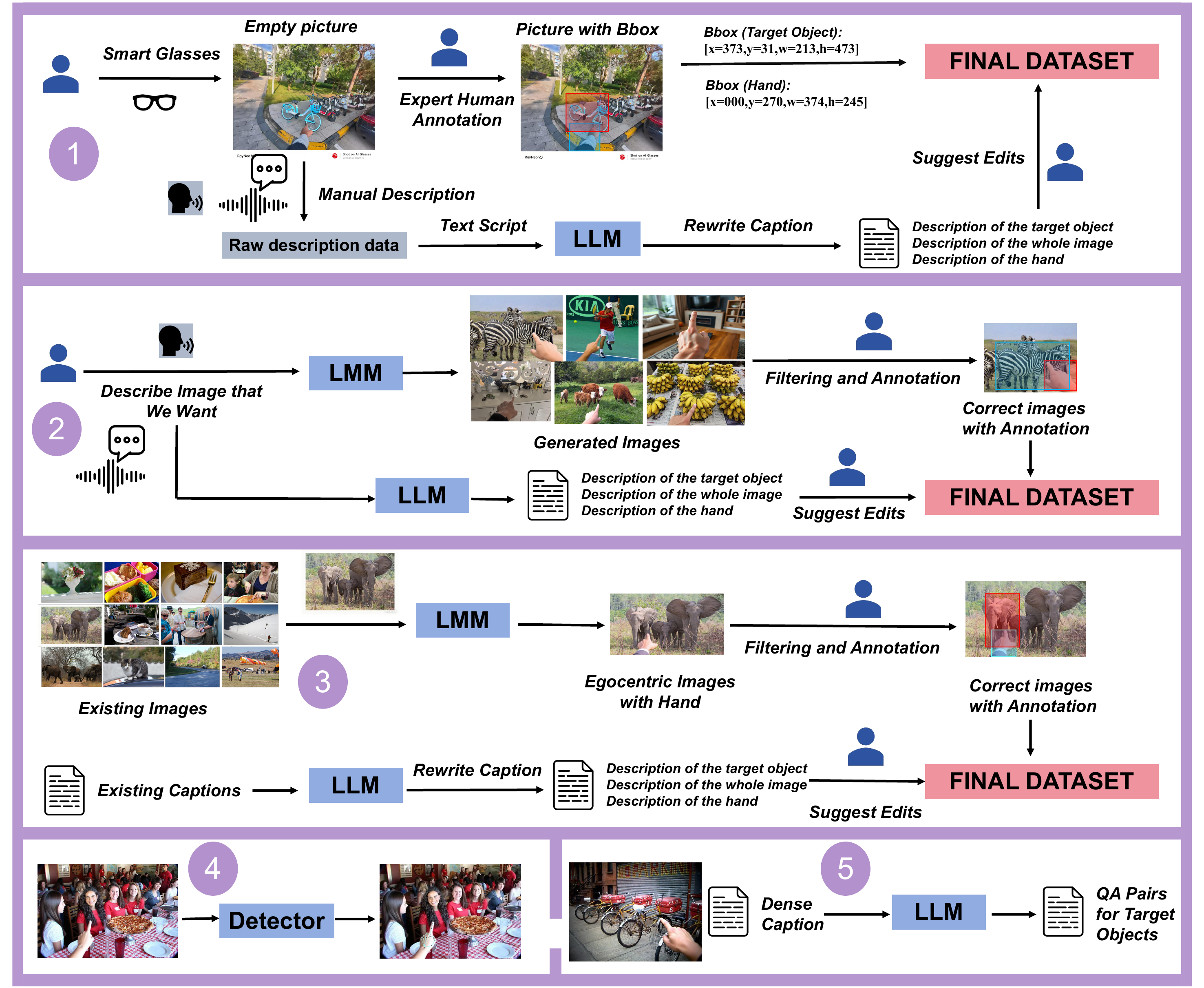}
    \caption{Overview of the Ego-Point Ground Dataset Collection and Multi-Stage Annotation Pipeline. Stages 1–3 illustrate the raw data acquisition and fundamental annotation workflow. Stage 4 details the extraction of 2D keypoint pose information. Subsequently, Stage 5 focuses on generating high-quality Visual Question-Answering (VQA) pairs for the target objects.}
    \label{fig:dataset collection}
\end{figure*}
We adopt a two-stage splitting strategy. First, under the mixed split, all data sources are combined and randomly partitioned into training, validation, and test sets at a $7:2:1$ ratio. Second, to simulate real-world deployment, we employ a domain-adaptive split: real-world data captured by smart glasses is reserved exclusively for testing, while synthetic and generative data are used for training. Specifically, real-captured, synthetically composed, and LLM-generated samples number $6{,}035$, $5{,}976$, and $3{,}327$, accounting for $39.35\%$, $38.96\%$, and $21.69\%$ of the dataset, respectively.

Category distribution reveals that person dominates at $47.1\%$, reflecting the inherent nature of egocentric perception—interactions between the wearer and other individuals constitute the most frequent and fundamental scenario. The remaining $19$ major categories are relatively balanced (each ranging from $2.2\%$ to $4.3\%$), as shown in Figure \ref{fig:statistic of dataset-word cloud}.

Data collection involved $17$ participants with diverse demographic attributes (gender, skin tone, age) across varied environments and time periods, ensuring comprehensive representation in both subject and environmental diversity.

\subsubsection{Cross-Modal Consistency Analysis}
We assess the cross-modal consistency between images and captions by randomly sampling $N_I$ images and $N_C$ captions and calculating the cosine similarity between their CLIP-generated \cite{radford2021learningclip} visual ($v_i$) and textual ($t_j$) embeddings:
\begin{equation}
    \text{sim}_{ij} = \frac{v_i \cdot t_j}{|v_i| |t_j|}.
\end{equation}
As illustrated in Figure \ref{fig:annotation analyze-heatmap and point}(a), the image-text similarity heatmap exhibits a pronounced diagonal dominance. This robust alignment confirms the precise semantic correspondence of the image-text pairs within our dataset, thereby validating its high annotation quality and consistency.
\subsubsection{Caption Semantic Diversity}
\noindent\textbf{Lexical Diversity:} Figure \ref{fig:statistic of dataset-word cloud}(c) shows that the dataset emphasizes interactive elements. High-frequency terms include common entities (e.g., "object", "person"), spatial indicators (e.g., "frame", "environment"), and dynamic interaction words (e.g., "pointing", "hand") paired with spatial terms (e.g., "right", "left"). This highlights the dataset's focus on context and dynamic egocentric behaviors.

\noindent\textbf{Semantic Uniformity:} To assess semantic diversity, we employ t-SNE \cite{linderman2017efficient-SNE} to project caption vector representations $t_i \in \mathbb{R}^d$ into a 2D scatter plot, as shown in Figure \ref{fig:annotation analyze-heatmap and point}(b). The results demonstrate significant spatial dispersion for most category captions, confirming high semantic diversity. A few categories exhibit tighter clustering, indicating higher linguistic consistency. The dataset's descriptive richness is supported by its broad spatial and categorical diversity.

\begin{figure*}
    \centering
    \includegraphics[width=\linewidth]{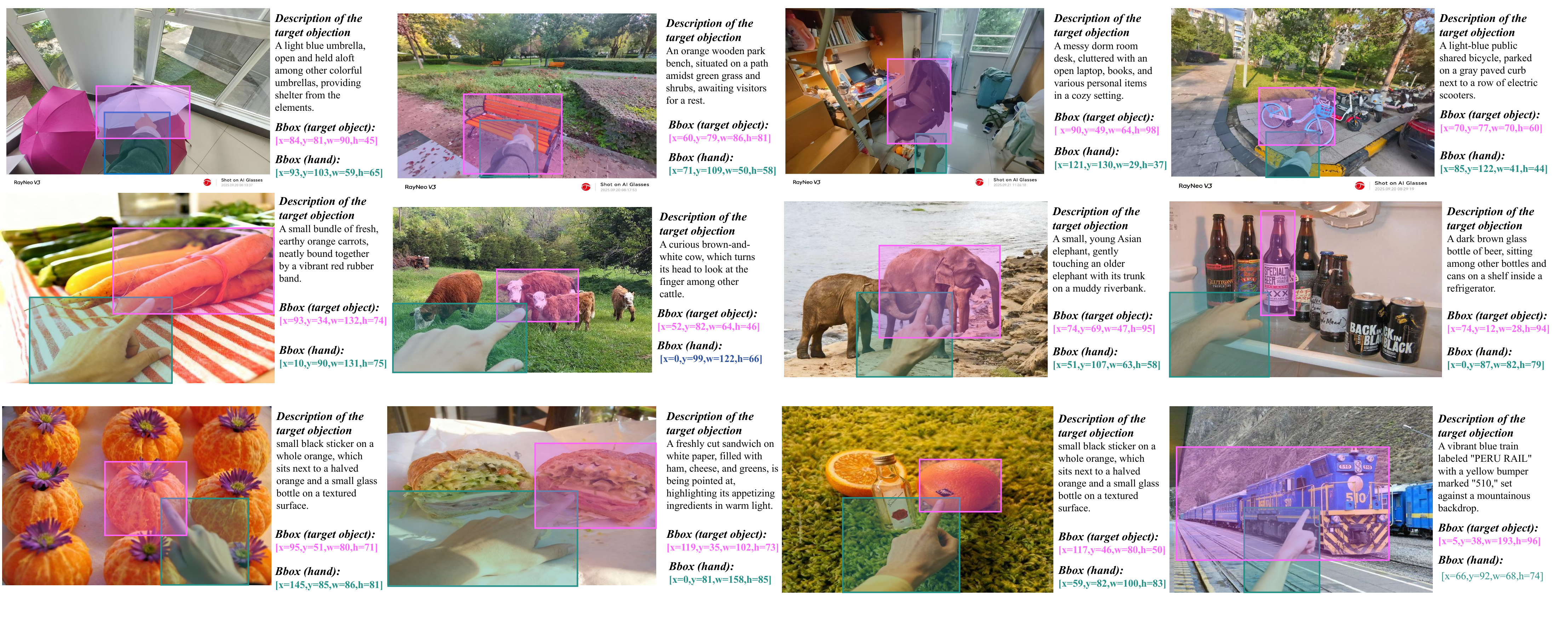}
    \caption{Image Examples of Ego-Point Ground Dataset. We showcase example images and partial annotations from the dataset, including bounding boxes for the hand and the target object, as well as the corresponding caption for the target object. The first row specifically illustrates data collected from our real-world captures.}
    \label{fig:dataset examples}
\end{figure*}


\subsection{Data Collection Pipeline}
The EgoPoint-Ground Dataset uses a Hybrid Data Generation Paradigm to blend real-world diversity with the efficiency of synthetic data, as shown in Figure \ref{fig:dataset collection}. It combines three main data sources, which are processed and annotated using a dedicated LMM-Aided Synthesis and Editing Pipeline \cite{gao2025seedream,yang2025qwen3}.
A partial example of the dataset is shown in Figure \ref{fig:dataset examples}.

\subsubsection{Real-World Data}
The first source ensures high fidelity by capturing real-world scenarios, We demonstrate this process in Stage 1 of Figure \ref{fig:dataset collection}. Researchers wearing smart glasses recorded egocentric images of hands pointing at objects in various environments (e.g., schools, supermarkets, homes). Data collectors provided detailed descriptions via voice, which were transcribed and initially drafted by an LLM. Human annotators then reviewed and corrected the LLM-generated text, and manually annotated pixel-accurate bounding boxes for the target object and hand region. This process produced the core caption components: image description, object referent, and hand description.

\subsubsection{Synthetic Generation}
The second component utilizes LMMs for directed image synthesis to efficiently generate challenging referential scenarios that are difficult to capture in the real world, such as complex occlusions or specific lighting conditions. The data collection procedure for this section is shown in Stage 2 of Figure \ref{fig:dataset collection}. Researchers manually filtered the synthetic Egocentric images to remove unqualified samples. For captioning, participants provided voice descriptions, which were transcribed by an LLM and then subjected to rigorous human review and correction. Target bboxes were manually labeled for precise localization.
\begin{figure*}
    \centering
    \includegraphics[width=\linewidth]{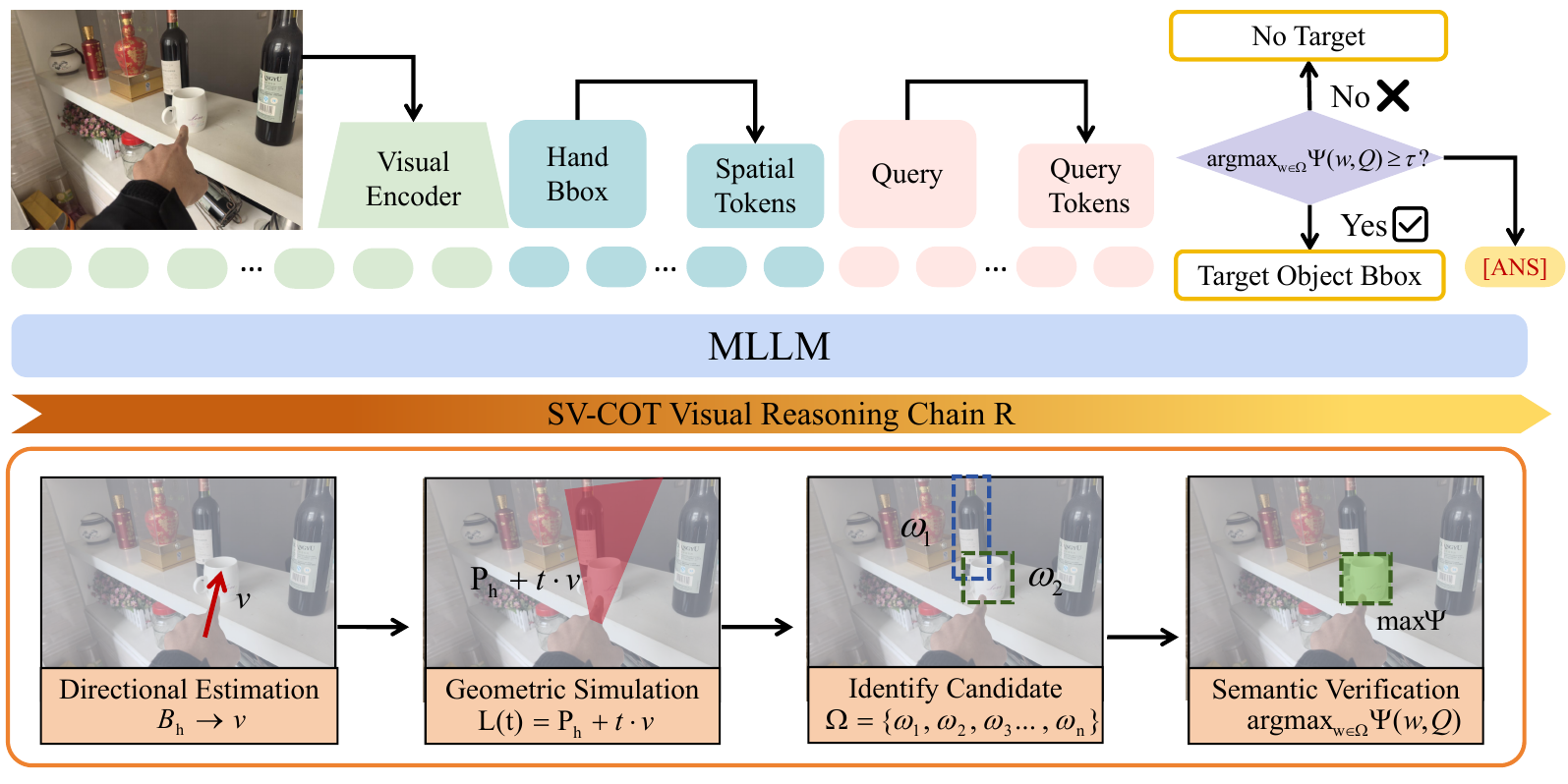}
    \caption{\textbf{Architectural overview of the SV-CoT framework.} The egocentric hand bounding box $\mathcal{B}_{hand}^{2D}$ is aligned into discrete spatial anchors $\mathcal{T}_{pos}$. Zero-shot grounding is reformulated as a latent reasoning chain $\mathcal{R}$ that sequentially infers: (1) a deictic directional primitive $\vec{v}$; (2) a virtual ray $\mathcal{L}(t)$ to spatially prune distractors into a candidate set $\Omega$; and (3) a multi-modal semantic verification score $\Psi$ to either resolve the target $BBox_{\mathcal{O}}$ or explicitly reject it via threshold $\tau$.}
    \label{fig:baseline model}
\end{figure*}

\subsubsection{Image Editing}
To efficiently scale the dataset and facilitate domain migration, we employed LMM-aided image editing, as shown in Figure \ref{fig:dataset collection}, stage 3. We selected third-person perspective images from existing datasets \cite{yu2016refcoco,lin2014microsoftcoco} and guided a Multi-modal Large Model (LMM) with Egocentric-specific prompts to edit them into egocentric scenes featuring hand pointing. The edited images underwent stringent human screening and secondary annotation to correct LMM-induced pointing inaccuracies. An LLM concurrently structured the captions into three parts: image description, object referent, and hand description.

\subsubsection{Auxiliary Annotation and Quality Control}
To enrich the dataset's dimensions and potential applications, we generated auxiliary annotations (Stage 4 and Stage 5 in Figure \ref{fig:dataset collection}). An advanced pose estimation model was used to process hand regions in the images, extracting keypoints for Hand Pose annotations, which were refined through human review. Simultaneously, we leveraged the fine-grained captions and an LLM to construct challenging referential QA Pairs.

\subsection{Ethical Considerations}

The EgoPoint-Ground dataset was collected following strict ethical and privacy guidelines. Participants gave written consent, understanding the data's use and their privacy rights. All images and metadata were de-identified to protect Personally Identifiable Information.

Access is limited to certified researchers through a formal licensing system, with clear usage rules and penalties for violations. We ensure full compliance with ethical standards and prioritize participant privacy.
\begin{table*}[t]
    \centering
    \caption{{Quantitative results of state-of-the-art models on the D-VQA task}, evaluated across both {Hybrid} (mixed synthetic and captured) and {Real-World} (naturally captured) datasets. Shaded rows categorize the models into distinct performance tiers, while the {Human Study} establishes the theoretical upper bound.}
    \label{tab:2-EDG-tier-organized}
    
    \setlength{\tabcolsep}{3pt} 
    \renewcommand{\arraystretch}{1.3} 
    \footnotesize 
    
    \begin{tabularx}{\textwidth}{@{} l *{6}{>{\centering\arraybackslash}X} @{}} 
        \toprule
        \multirow{2}{*}{\textbf{Model}} & \multicolumn{3}{c}{\textbf{EDG --- Hybrid}} & \multicolumn{3}{c}{\textbf{EDG --- Real-World}} \\
        \cmidrule(lr){2-4} \cmidrule(lr){5-7}
        & \textbf{$\mathcal{P}@0.3\uparrow$} & \textbf{$\mathcal{P}@0.5\uparrow$} & \textbf{$\mathcal{P}@0.7\uparrow$} & \textbf{$\mathcal{P}@0.3\uparrow$} & \textbf{$\mathcal{P}@0.5\uparrow$} & \textbf{$\mathcal{P}@0.7\uparrow$} \\
        \midrule
        
        Qwen3-VL-4B \cite{yang2025qwen3}           & 0.743 & 0.702 & 0.622 & 0.747 & 0.699 & 0.616 \\
        Qwen2.5-VL-3B \cite{wang2024qwen2-vl}      & 0.731 & 0.671 & 0.592 & 0.738 & 0.632 & 0.488 \\
        Qwen3-VL-8B \cite{yang2025qwen3}           & 0.726 & 0.676 & 0.598 & 0.789 & 0.701 & 0.593 \\
        Qwen2.5-VL-7B \cite{wang2024qwen2-vl}      & 0.721 & 0.653 & 0.563 & 0.776 & 0.688 & 0.533 \\
        \addlinespace[4pt]

        LLaVA-OneVision-4B \cite{an2025llavaoneversion}    & 0.679 & 0.475 & 0.145 & 0.654 & 0.431 & 0.129 \\
        InternVL-3.5-8B \cite{wang2025internvl3}           & 0.554 & 0.413 & 0.270 & 0.592 & 0.427 & 0.214 \\
        GroundingGPT \cite{li2024groundinggpt}               & 0.447 & 0.358 & 0.219 & 0.354 & 0.259 & 0.127 \\
        DeepSeek-VL2 \cite{deepseek-vl2_wu2024deepseek}       & 0.344 & 0.293 & 0.259 & 0.364 & 0.296 & 0.245 \\
        \addlinespace[4pt]

        LLaVA-v1.5-13B \cite{llava-1.5-13b_liu2024improved} & 0.338 & 0.198 & 0.079 & 0.274 & 0.128 & 0.042 \\
        Ferret-7B \cite{you2023ferret}                   & 0.282 & 0.221 & 0.181 & 0.258 & 0.203 & 0.160 \\
        Florence2-Large \cite{xiao2024florence}               & 0.241 & 0.188 & 0.149 & 0.234 & 0.193 & 0.153 \\
        Florence2-Base \cite{xiao2024florence}                & 0.149 & 0.110 & 0.080 & 0.141 & 0.121 & 0.075 \\
        \midrule
        
        \rowcolor{gray!15}
        \textbf{SV-CoT [Qwen3-vl] (Ours)} & \textbf{0.795} & \textbf{0.756} & \textbf{0.718} & \textbf{0.820} & \textbf{0.818} & \textbf{0.727} \\
        \rowcolor{blue!5} 
        \textbf{Human Study}   & 0.979 & 0.914 & 0.846 & 0.994 & 0.918 & 0.874 \\
        \bottomrule
    \end{tabularx}
\end{table*}
\vspace{-8pt}
\section{Task Setting}
In this section, we establish a comprehensive benchmark for Egocentric Deictic Visual Grounding. By introducing high-quality annotations alongside a robust baseline framework, we provide a solid groundwork to catalyze future research in this novel domain. Leveraging the newly constructed EgoPoint-Ground dataset, we formally define the following evaluation tasks:
\subsection{Egocentric Deictic Visual Grounding (EDG)}
\label{subsec:task-EDG}
\vspace{-8pt}

\noindent\textbf{Motivation}: In real-world embodied interactions, deictic gestures and natural language exhibit strong synergistic complementarity. While physical pointing provides an immediate spatial prior, linguistic descriptions supply essential semantic attributes. Because neither modality is independently sufficient to resolve complex spatial or semantic ambiguities, we formulate the Egocentric Deictic Grounding (EDG) task. This benchmark challenges models to seamlessly integrate static hand-pointing cues with linguistic context to achieve robust cross-modal localization.

\noindent\textbf{Task Definition}: Formally, the EDG task requires the precise spatial localization of a target object $\mathcal{O}$ through the fusion of visual and linguistic information. The input consists of an egocentric image $\mathcal{I}$, the 2D bounding box of the pointing hand $\mathcal{B}_{hand}^{2D}$, and an underspecified linguistic referent $\mathcal{D}_{\mathcal{O}}$ (e.g., "this object" or "that boy"). Notably, $\mathcal{D}_{\mathcal{O}}$ identifies the target's category or intent but lacks unique spatial discriminability.
Input: $(\mathcal{I}, \mathcal{B}_{hand}^{2D}, \mathcal{D}_{\mathcal{O}})$; Output: The 2D bounding box of the target object, $BBox_{\mathcal{O}}$.

\subsection{Deictic Visual Question Answering (D-VQA)} 
\label{subsec:task-D-VQA}
\noindent\textbf{Motivation:} The ultimate goal of advanced embodied agents is to achieve meaningful, situated interaction and manipulation. Following successful localization, high-level reasoning about the target object in the environment is often required to determine subsequent actions and intent. The D-VQA task aims to evaluate a model’s ability to perform high-level Situated Reasoning on the target object after successfully parsing the hand pointing intent and localization, thereby supporting decision-making and action planning.
\begin{table*}[t]
    \centering
    \caption{Performance of state-of-the-art models on the D-VQA task across \textbf{Hybrid} (mixed synthetic/captured) and \textbf{Real-World} (exclusively captured) datasets. The "Human Study" denotes the evaluation results from human participants.}
    \label{tab:deictic_vqa_final_style}
    
    \setlength{\tabcolsep}{4pt} 
    \renewcommand{\arraystretch}{1.3} 
    \footnotesize 
    
    \begin{tabularx}{\textwidth}{@{} l *{6}{>{\centering\arraybackslash}X} @{}} 
        \toprule
        \multirow{2}{*}{\textbf{Model}} & \multicolumn{3}{c}{\textbf{D-VQA (Hybrid)}} & \multicolumn{3}{c}{\textbf{D-VQA (Real-World)}} \\
        \cmidrule(lr){2-4} \cmidrule(lr){5-7}
        & \textbf{Acc.}$\uparrow$ & \textbf{M-F1}$\uparrow$ & \textbf{Rec.}$\uparrow$ & \textbf{Acc.}$\uparrow$ & \textbf{M-F1}$\uparrow$ & \textbf{Rec.}$\uparrow$ \\
        \midrule
        
        LLaVA-OneVision-8B \cite{an2025llavaoneversion} & 0.531 & 0.519 & 0.548 & 0.544 & 0.525 & 0.526 \\
        LLaVA-OneVision-4B \cite{an2025llavaoneversion} & 0.510 & 0.535 & 0.547 & 0.521 & 0.476 & 0.487 \\
        Qwen3-VL-8B \cite{yang2025qwen3}                & 0.494 & 0.484 & 0.490 & 0.517 & 0.510 & 0.503 \\
        Qwen3-VL-4B \cite{yang2025qwen3}                & 0.471 & 0.488 & 0.495 & 0.456 & 0.507 & 0.489 \\
        \addlinespace[4pt]
        
        Qwen2.5-VL-7B \cite{wang2024qwen2-vl}           & 0.457 & 0.427 & 0.445 & 0.444 & 0.428 & 0.411 \\
        Qwen2.5-VL-3B \cite{wang2024qwen2-vl}           & 0.456 & 0.461 & 0.470 & 0.441 & 0.527 & 0.523 \\
        DeepSeek-VL2-Small \cite{deepseek-vl2_wu2024deepseek} & 0.413 & 0.459 & 0.455 & 0.457 & 0.523 & 0.506 \\
        DeepSeek-VL2 \cite{deepseek-vl2_wu2024deepseek}       & 0.412 & 0.461 & 0.473 & 0.393 & 0.397 & 0.390 \\
        \addlinespace[4pt]
        
        InternVL-3.5-4B \cite{wang2025internvl3}        & 0.445 & 0.461 & 0.445 & 0.433 & 0.368 & 0.367 \\
        InternVL-3.5-8B \cite{wang2025internvl3}        & 0.316 & 0.392 & 0.381 & 0.286 & 0.398 & 0.380 \\
        GroundingGPT \cite{li2024groundinggpt}          & 0.334 & 0.363 & 0.364 & 0.392 & 0.352 & 0.350 \\
        LLaVA-v1.5-13B \cite{llava-1.5-13b_liu2024improved} & 0.267 & 0.249 & 0.245 & 0.147 & 0.150 & 0.137 \\
        \midrule
        
        \rowcolor{gray!15} 
        
        \rowcolor{blue!5} 
        \textbf{Human Study}   &0.811  &0.631  &0.653  &0.877  &0.756  &0.753  \\
        \bottomrule
    \end{tabularx}
\end{table*}
\noindent\textbf{Task Definition: } Given the image $\mathcal{I}$, the hand's 2D bbox $\mathcal{B}_{hand}^{2D}$, and a specific question about the referred object $\mathcal{Q}$ (eg. "What is this?", "What is the boy wearing? "), the model must provide the answer.
Input: $(\mathcal{I}, \mathcal{B}_{hand}^{2D}, \mathcal{Q})$; 
Output: The answer to the question, $\mathcal{A}$.

\subsection{Language-Only Grounding(D-REC)}
\label{subsec:task-REC}

\noindent\textbf{Motivation:} The Language-Only Grounding task follows the paradigm of traditional Referring Expression Comprehension (REC) but is specifically tailored for egocentric deictic scenarios. The motivation is twofold: first, to quantify the model's independent capability to interpret referential linguistic cues (e.g., "this", "this boy") within a first-person visual context; and second, to serve as a rigorous ablation baseline for evaluating the performance gain of multi-modal fusion (as in EDG) over a purely linguistic-driven approach on the same dataset.

\noindent\textbf{Task Definition:}
Input: $(\mathcal{I}, \mathcal{D}_{\mathcal{O}})$;
Output: $BBox_{\mathcal{O}}$.

\subsection{Physical Pointing-Only Grounding (POG)}
\label{subsec:task-POG}

\noindent\textbf{Motivation:} This task aims to quantify the independent referencing capability of the hand-pointing pose. By providing only the 2D pose, we evaluate the model’s ability to infer spatial pointing intent purely from the static hand configuration. This forms another EDG ablation baseline.

\noindent\textbf{Task Definition: }
Input: $(\mathcal{I}, \mathcal{B}_{hand}^{2D})$;
Output: $BBox_{\mathcal{O}}$.



\section{Experiment}
\subsection{Baseline Method}
\label{sec:method}

To bridge the granularity gap between continuous visual coordinates and the discrete linguistic vocabulary of MLLMs, we adopt a coordinate-to-token alignment protocol. Given an egocentric image $\mathcal{I}$ and the referrer's hand bounding box $\mathcal{B}_{hand}^{2D} = [x, y, w, h]$, we apply a spatial quantization function $\phi(\cdot)$ to map each coordinate into $N$ discrete spatial bins (where $N=1000$):
\begin{equation}
    \mathcal{T}_{pos} = \left\{ \langle \text{bin}_c \rangle \mid c \in \mathcal{B}_{hand}^{2D}, \phi(c) = \left\lfloor \frac{c}{L} \cdot (N-1) \right\rfloor \right\}
\end{equation}
These quantized tokens $\mathcal{T}_{pos}$ serve as \textbf{Spatial Anchors}, grounding the LLM's attention to the specific ego-gestural region within the multimodal input stream.

\subsubsection{SV-CoT: Spatial-Reasoning-based Visual Chain-of-Thought}
Instead of treating grounding as a direct regression task, we reformulate it as a structured inference process. We present \textbf{SV-CoT}, a zero-shot inference framework instantiated on Qwen3-VL \cite{yang2025qwen3}. As depicted in Figure \ref{fig:baseline model}, this baseline systematically decomposes the complex deictic intent into a sequential latent reasoning chain $\mathcal{R}$. Formally, the target distribution $P(BBox_{\mathcal{O}} \mid \mathcal{I}, \mathcal{B}_{hand}^{2D}, \mathcal{Q})$ is modeled by marginalizing over a sequence of geometric rationales:
\begin{equation}
    P(BBox_{\mathcal{O}} \mid \mathcal{X}_{in}) = \sum_{\mathcal{R}} P(BBox_{\mathcal{O}} \mid \mathcal{R}, \mathcal{X}_{in}) P(\mathcal{R} \mid \mathcal{X}_{in})
\end{equation}
where $\mathcal{X}_{in} = (\mathcal{I}, \mathcal{B}_{hand}^{2D}, \mathcal{Q})$ denotes the multimodal input context. The reasoning variable $\mathcal{R} = \{r_{dir}, r_{path}, r_{ref}\}$ is instantiated as a causal chain of linguistic tokens, mimicking human-like spatial deduction:

\paragraph{1. Deictic Intent Parsing ($r_{dir}$):} 
The model first performs an \textit{anatomical-to-spatial} projection. By analyzing the orientation of the hand within the anchor $\mathcal{B}_{hand}^{2D}$, the MLLM instantiates a directional primitive $\vec{v} \in \mathbb{R}^2$, effectively transforming local gesture posture into a global pointing prior.

\paragraph{2. Geometric Trajectory Simulation ($r_{path}$):} 
This primitive defines a virtual ray $\mathcal{L}(t) = \mathcal{P}_{hand} + t \cdot \vec{v}$, where $\mathcal{P}_{hand}$ denotes the spatial centroid of $\mathcal{B}_{hand}^{2D}$. The model executes a \textbf{spatial pruning} operation, identifying a candidate set $\Omega = \{ \omega_1, \dots, \omega_n \}$ of visual regions that satisfy the intersection constraint $\omega_i \cap \mathcal{L}(t) \neq \emptyset$. This effectively filters spatial distractors from cluttered egocentric scenes.

\paragraph{3. Spatio-semantic Verification ($r_{ref}$):} 
The final phase involves a multi-modal alignment check. The model resolves the target $BBox_{\mathcal{O}}$ by evaluating the semantic consistency $\Psi$ between the linguistic query $\mathcal{Q}$ and the pruned candidates $\Omega$:
\begin{equation}
    BBox_{\mathcal{O}} = \begin{cases} 
    \arg\max_{\omega \in \Omega} \Psi(\omega, \mathcal{Q}) & \text{if } \max_{\omega \in \Omega} \Psi(\omega, \mathcal{Q}) \geq \tau \\
    \varnothing & \text{otherwise}
    \end{cases}
\end{equation}
where $\tau$ is a confidence threshold for \textbf{explicit rejection}. By parameterizing the zero-shot inference as $\mathcal{Y} = [\mathcal{R}, BBox_{\mathcal{O}}]$, SV-CoT enforces a structural constraint that mitigates spatial hallucination and ensures referential consistency.

\begin{table*}[!t]
    \centering
    \caption{Performance comparison on conventional REC and dedicated POG tasks. Absolute scores are reported with relative gains/losses compared to the baseline (Inputs are provided without either spatial or linguistic cues) in parentheses.}
    \label{tab:REC_and_POG_refined}
    
    \setlength{\tabcolsep}{2.5pt} 
    \renewcommand{\arraystretch}{1.3} 
    \footnotesize 
    
    \begin{tabularx}{\textwidth}{@{} l *{6}{>{\centering\arraybackslash}X} @{}} 
        \toprule
        \multirow{2}{*}{\textbf{Model}} & \multicolumn{3}{c}{\textbf{D-REC}} & \multicolumn{3}{c}{\textbf{POG}} \\
        \cmidrule(lr){2-4} \cmidrule(lr){5-7}
        & \textbf{$\mathcal{P}@0.3\uparrow$} & \textbf{$\mathcal{P}@0.5\uparrow$} & \textbf{$\mathcal{P}@0.7\uparrow$} & \textbf{$\mathcal{P}@0.3\uparrow$} & \textbf{$\mathcal{P}@0.5\uparrow$} & \textbf{$\mathcal{P}@0.7\uparrow$} \\
        \midrule
        
        Qwen3-VL-4B \cite{yang2025qwen3}           & 0.737 \pos{0.087} & 0.684 \pos{0.118} & 0.605 \pos{0.123} & 0.671 \pos{0.021} & 0.612 \pos{0.046} & 0.532 \pos{0.050} \\
        Qwen3-VL-8B \cite{yang2025qwen3}           & 0.701 \pos{0.111} & 0.648 \pos{0.087} & 0.561 \pos{0.077} & 0.653 \pos{0.063} & 0.595 \pos{0.034} & 0.514 \pos{0.030} \\
        Qwen2.5-VL-7B \cite{wang2024qwen2-vl}      & 0.703 \pos{0.190} & 0.638 \pos{0.160} & 0.541 \pos{0.125} & 0.670 \pos{0.157} & 0.611 \pos{0.143} & 0.515 \pos{0.099} \\
        Qwen2.5-VL-3B \cite{wang2024qwen2-vl}      & 0.665 \pos{0.169} & 0.592 \pos{0.167} & 0.491 \pos{0.139} & 0.575 \pos{0.079} & 0.501 \pos{0.076} & 0.407 \pos{0.055} \\
        \addlinespace[4pt]

        LLaVA-OneVision-4B \cite{an2025llavaoneversion}    & 0.586 \pos{0.194} & 0.347 \pos{0.186} & 0.088 \pos{0.071} & 0.525 \pos{0.133} & 0.292 \pos{0.131} & 0.065 \pos{0.048} \\
        LLaVA-OneVision-8B \cite{an2025llavaoneversion}    & 0.505 \pos{0.167} & 0.258 \pos{0.110} & 0.049 \pos{0.027} & 0.403 \pos{0.065} & 0.181 \pos{0.033} & 0.025 \pos{0.003} \\
        InternVL-3.5-4B \cite{wang2025internvl3}           & 0.535 \pos{0.149} & 0.402 \pos{0.135} & 0.272 \pos{0.092} & 0.416 \pos{0.030} & 0.305 \pos{0.038} & 0.203 \pos{0.023} \\
        InternVL-3.5-8B \cite{wang2025internvl3}           & 0.456 \pos{0.066} & 0.325 \pos{0.045} & 0.210 \pos{0.006} & 0.497 \pos{0.107} & 0.311 \pos{0.031} & 0.178 \negv{0.026} \\
        DeepSeek-VL2-Small \cite{deepseek-vl2_wu2024deepseek} & 0.443 \negv{0.001} & 0.393 \negv{0.005} & 0.341 \pos{0.023} & 0.375 \negv{0.069} & 0.318 \negv{0.080} & 0.274 \negv{0.044} \\
        \addlinespace[4pt]

        Ferret-7B \cite{you2023ferret}                   & 0.428 \pos{0.029} & 0.350 \pos{0.015} & 0.261 \negv{0.001} & 0.413 \pos{0.014} & 0.353 \pos{0.018} & 0.281 \pos{0.019} \\
        GroundingGPT \cite{li2024groundinggpt}               & 0.378 \negv{0.149} & 0.261 \negv{0.169} & 0.129 \negv{0.162} & 0.536 \pos{0.009} & 0.443 \pos{0.013} & 0.305 \pos{0.014} \\

        \midrule
        \rowcolor{gray!5}
        \textbf{SV-CoT [Qwen3-vl] (Ours)} & \textbf{0.814} & \textbf{0.750} & \textbf{0.694} & \textbf{0.694} & \textbf{0.639} & \textbf{0.556} \\
        \rowcolor{blue!3} 
        \textbf{Human Study}   & 0.952 & 0.926 & 0.819 & 0.956 & 0.916 & 0.806 \\
        \bottomrule
    \end{tabularx}
    \vspace{-5pt}
\end{table*}
\vspace{-10pt}
\subsection{Evaluation Metrics}
We evaluate the proposed benchmarks using the following metrics:

\noindent\textbf{Localization Metrics:} For EDG, D-REC, and POG tasks, we adopt {Precision@$\tau$} as the core metric, which measures the proportion of samples where the Intersection over Union (IoU) between the predicted and ground-truth bounding boxes exceeds threshold $\tau$. We report results at $\tau \in \{0.3, 0.5, 0.7\}$ to reflect varying localization precision.

\noindent\textbf{Classification Metrics:} For the D-VQA task, we employ {Accuracy (Acc.)}, {Macro F1-score}, and {Macro Recall} to assess the correctness of answers. Macro-averaging is specifically used to ensure a fair evaluation across the class-imbalanced distribution of our dataset.
\subsection{Experimental Results}
\subsubsection{Results on Egocentric Deictic Grounding}
We conduct a comprehensive evaluation of state-of-the-art MLLMs and VG models on the EDG task across both {hybrid} and {real-world} scenarios (see Table \ref{tab:2-EDG-tier-organized}). The results reveal a distinct performance hierarchy: the Qwen3 and Qwen2.5 series lead the field due to their superior vision-language alignment (e.g., Qwen3-VL-8B achieves $ \mathcal{P}@0.5$ of $0.701$ on {real-world}), whereas foundational models like Florence2 and LLaVA-v1.5 struggle, with accuracies consistently below $0.34$. Notably, while baseline models exhibit a sharp performance decline at higher precision thresholds ($\mathcal{P}@0.7$), highlighting the significant challenge of fine-grained gesture-to-object localization, our method demonstrates remarkable resilience.
Specifically, a substantial gap of over $26\%$ persists between the top-performing baseline and the \textbf{Human Study} upper bound ($\mathcal{P}@0.5 \ge 0.914$), suggesting that general-purpose MLLMs have yet to master complex "hand-object" spatial logic.
Our proposed baseline, \textbf{SV-CoT}, explicitly guides spatial reasoning via a Visual Chain-of-Thought, achieving an impressive absolute improvement of $11.7\%$ ($0.818$ vs. $0.701$) in {real-world} and $5.4\%$ ($0.756$ vs. $0.702$) in {hybrid} scenarios at $\mathcal{P}@0.5$ over the strongest open-source baselines. Furthermore, SV-CoT effectively maintains a high accuracy of over $0.71$ even at the strict $\mathcal{P}@0.7$ threshold, significantly mitigating semantic ambiguity in deictic references. Interestingly, most baseline models perform slightly better in real-world than in hybrid scenarios. This discrepancy arises because the hybrid set serves as a rigorous stress test, incorporating a higher density of synthetic noise, extreme viewpoint augmentations, and occlusions that surpass the distributional complexity of natural captures.
\vspace{-10pt}
\subsubsection{Results on Deictic Visual Question Answering}

The D-VQA task requires models to establish deep semantic correlations between hand-pointing gestures and specific linguistic queries within complex visual scenes. As shown in Table \ref{tab:deictic_vqa_final_style}, LLaVA-OneVision and the Qwen3-VL series demonstrate competitive performance, with LLaVA-OneVision-8B achieving the highest accuracy of 0.544 in real-world scenarios. In contrast, foundational or smaller-scale models (e.g., LLaVA-v1.5) struggle with such gesture-based visual reasoning.

Despite progress by top-tier models, a performance gap of approximately 30\% remains relative to the \textbf{Human Study} upper bound (Acc. $> 0.81$). This gap underscores the inherent difficulty for general-purpose MLLMs in precisely mapping spatial coordinates (Hand Bbox) to semantic question-answering, a challenge we define as deictic alignment failure. 
\vspace{-10pt}
\subsubsection{Ablation Studies: Linguistic vs. Geometric Cues}

D-REC and POG evaluate referential understanding using isolated linguistic or physical cues (Table \ref{tab:REC_and_POG_refined}). Compared to the multimodal EDG task, all models show significant performance drops, confirming the severe referential ambiguity of unimodal inputs. While Qwen3 maintains a robust $\mathcal{P}@0.5$ of $0.684$ in D-REC, POG results expose a critical deficiency in geometric understanding. Models generally perform worse on POG, collapsing at high-precision thresholds (e.g., LLaVA-OneVision-8B drops to $0.025$ at $\mathcal{P}@0.7$). This indicates that current MLLMs fail to map the spatial trajectory from "fingertip" to "target" without linguistic aid, revealing a massive gap in physical intent comprehension compared to human performance.

Our baseline, \textbf{SV-CoT}, effectively mitigates these restrictions via a Visual Chain-of-Thought. It achieves $0.639$ at $\mathcal{P}@0.5$ in POG (an absolute gain of $2.7\%$ over the strongest baseline) and $0.750$ in D-REC (a $6.6\%$ gain). This demonstrates that explicit gesture-object spatial modeling effectively overcomes the limitations of isolated geometric or linguistic perception.
\vspace{-10pt}
\section{Conclusion}


In this work, we formalized the egocentric deictic visual grounding task, directly addressing the inherent inadequacies of text-only queries in resolving physical interaction ambiguities. Our primary contribution is EgoPoint-Ground, a pioneering large-scale multimodal dataset comprising over 15,000 densely annotated samples that capture hand-gesture and linguistic co-reference. Upon this foundation, we established a rigorous benchmark evaluating a wide spectrum of state-of-the-art MLLMs and visual grounding architectures. Furthermore, we introduced SV-CoT, a robust baseline framework that synergizes gestural primitives and linguistic context through a Visual Chain-of-Thought reasoning mechanism. We will open-source our dataset and models to advance research in embodied AI, egocentric vision, and multimodal interaction.

\noindent\textbf{Limitations and Future Work}
While EgoPoint-Ground provides a robust baseline, it currently focuses on canonical index-finger pointing in static images, limiting the exploration of complex gestures and continuous spatiotemporal trajectories. Furthermore, extreme cases involving severe occlusion or highly distal targets remain challenging. To capture dynamic pointing intents, future work will expand this paradigm into the video domain.

\clearpage  


%
%
\bibliographystyle{splncs04}
\bibliography{main}
\clearpage
\section{Supplementary Material}
\section*{Related Work}
Perception from an egocentric perspective is crucial in Embodied AI \cite{pfeifer2004embodied,liu2025aligningembodied,duan2022surveyembodied,wang2023voyager}, robotics \cite{banyai2024robotics,wang2025largerobotics,team2025geminirobotics,garcia2007evolutionrobotics}, and Augmented Reality (AR) \cite{carmigniani2011augmented,arena2022overviewaugmented}. In egocentric interactions \cite{bansal2022egoprocess,grauman2022Ego4d,grauman2024Ego-exo4d}, the hand plays a key role not only in actions but also in communication and pointing \cite{chen2024largepointing,schauerte2010pointAT}. Pointing gestures are critical for providing spatial grounding for language \cite{hashi2024systematichand,narasimhaswamy2024handiffuser}. Prior work on hand-object interactions \cite{bansal2022egoprocess, grauman2022Ego4d, grauman2024Ego-exo4d} and datasets like YouRefit Pointing at Objects \cite{chen2021yourefit} have explored pointing gestures, while others like referAT \cite{2010referAt} and PointAt \cite{schauerte2010pointAT} integrate language for object reference tasks. However, these datasets either focus on long-term actions (e.g., “cooking”) without fine-grained instantaneous reference annotations or offer limited interaction cues \cite{hashi2024systematichand,zhao2022metaverse,guo2021humaninteraction}. Our dataset focuses on a key moment: an instant where an object is referenced by a hand gesture in the first-person view. True embodied interaction requires models to understand not just "what is happening," but "what is being referenced."

\subsection{Visual Grounding}
\label{subsec:related work-vg}
Visual Grounding \cite{zhang2018grounding,dai2024simvg,xiao2024oneref,zhuang2018parallel,xiao2024towardssurvey,anderson2018vision}, particularly Referring Expression Comprehension (REC) \cite{hu2017modeling,yu2018mattnet,yang2019dynamic,su2024scanformer,qiao2020referringsurvey}, aims to localize the region in an image corresponding to a natural language description.
\subsubsection{Traditional REC and Viewpoint Limitations}
Traditional REC tasks have been extensively studied under the third-person viewpoint. Classic datasets, such as RefCOCO \cite{yu2016refcoco,lin2014microsoftcoco}, RefCOCO+ \cite{mao2016refcoco+}, and RefCOCOg \cite{nagaraja2016refcocog}, have driven the evolution of models from CNN-based \cite{he2016deepresnet,hussain2023yolo,wang2025yoloe,cheng2024yoloworld,he2017maskecnn,ma2022understandingCNN} architectures to Transformers \cite{su2024scanformer,vaswani2017attentionisall}. Recently, open-vocabulary detectors like Grounding DINO \cite{liu2024groundingdino,zhang2022dino,oquab2023dinov2} and YOLO-World \cite{cheng2024yoloworld,wang2025yoloe,hussain2023yolo} have demonstrated strong performance on these benchmarks. However, these datasets and models represent a limited definition of visual perception \cite{efron1969perception,ito2024understandingperception}. They primarily stem from isolated captures in a "bystander" view, completely neglecting the presence of an interacting agent (e.g., a user), let alone non-verbal cues such as hand gestures and intentions.
\subsubsection{Towards Egocentric Visual Grounding}
Migrating the REC task to the egocentric viewpoint introduces unique challenges, such as severe object occlusion \cite{pepikj2013occlusion,chandel2015occlusion}, drastic viewpoint changes \cite{efron1969perception,ito2024understandingperception}, and visual interference caused by the hand itself \cite{narasimhaswamy2024handiffuser,liu2024deeppose,andriluka20142dpose}. Crucially, the hand also offers a key piece of structured non-verbal information. Distinct from prior research, our work explicitly treats the hand gesture as a signal, rather than noise, aiming to study the complex, tripartite relationship between language, gesture, and the visual target.
\subsection{MLLMs for REC}
\label{subsec:related work-MLLM}
In recent years, Multimodal Large Language Models (MLLMs) \cite{team2023gemini,liu2024deepseek,chuang2025meta,achiam2023gpt,zhang2024llava,touvron2023llama}, such as LLaVA \cite{zhang2024llava}, Ferret \cite{you2023ferret}, Qwen-VL \cite{bai2023qwen,yang2025qwen3}, and kosmos \cite{peng2023kosmos}, have significantly enhanced visual understanding and localization generalization, including zero-shot \cite{xian2017zero,bansal2018zero,pourpanah2022reviewzero} capabilities, through pre-training on vast image-text data. Models like Ferret \cite{you2023ferret} have demonstrated the potential to output bounding boxes directly, addressing the REC task. Nevertheless, their core framework remains primarily centered on (image, text) alignment. They lack an effective mechanism to integrate fine-grained \cite{chu2024fine,guo2024fine,liu2024focus}, structured non-verbal information, such as the 2D pose of a pointing gesture or precise physical spatial relationships \cite{liu2024deeppose,andriluka20142dpose}.
\begin{figure*}
    \centering
    \includegraphics[width=\linewidth]{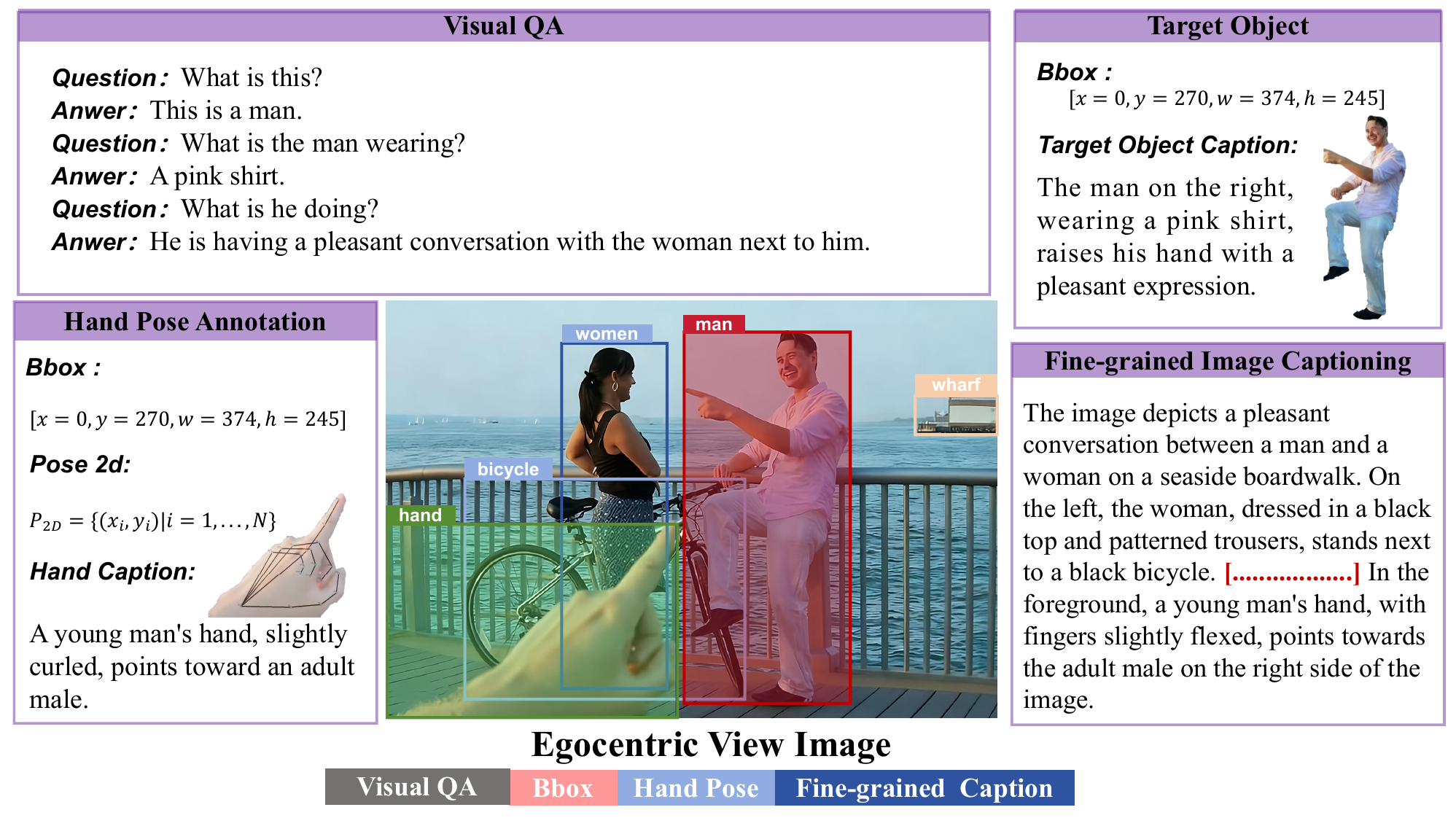}
    \caption{Overview of the Egopoint-ground Dataset and its Core Annotations. The Egopoint-ground dataset comprises first-person-perspective images captured via smart glasses, depicting a hand pointing towards a target object. The dataset features four primary annotations for each image: 1) Target Object Annotation: Includes the object's bounding box, textual description, and essential attributes. 2) Fine-grained Description: Detailed linguistic descriptions of the overall scene and the deictic (referential) interaction. 3) High-Quality Visual QA Pairs: Question-and-answer pairs focusing on the target object and the interaction intent. 4) First-Person Hand Pose and Action: Provides the hand's bounding box, 2D keypoint pose, and precise action description, serving as a key non-verbal spatial anchor.}
    \label{fig:dataset show supple}
\end{figure*}


\begin{figure*}[!t]
    \centering
    \includegraphics[width=\linewidth]{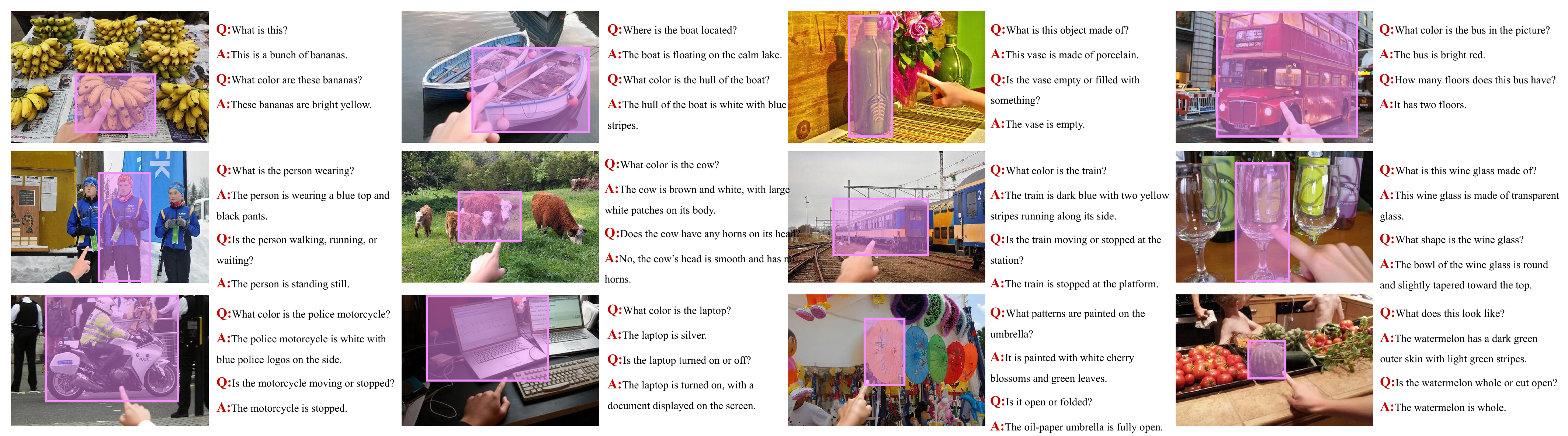}
    \caption{Example of the Visual Question Answering Section for Object Pointing, Showcased via Random Sampling.}
    \label{fig:QA supple}
\end{figure*}
\begin{figure*}[!t]
    \centering
    \includegraphics[width=\linewidth]{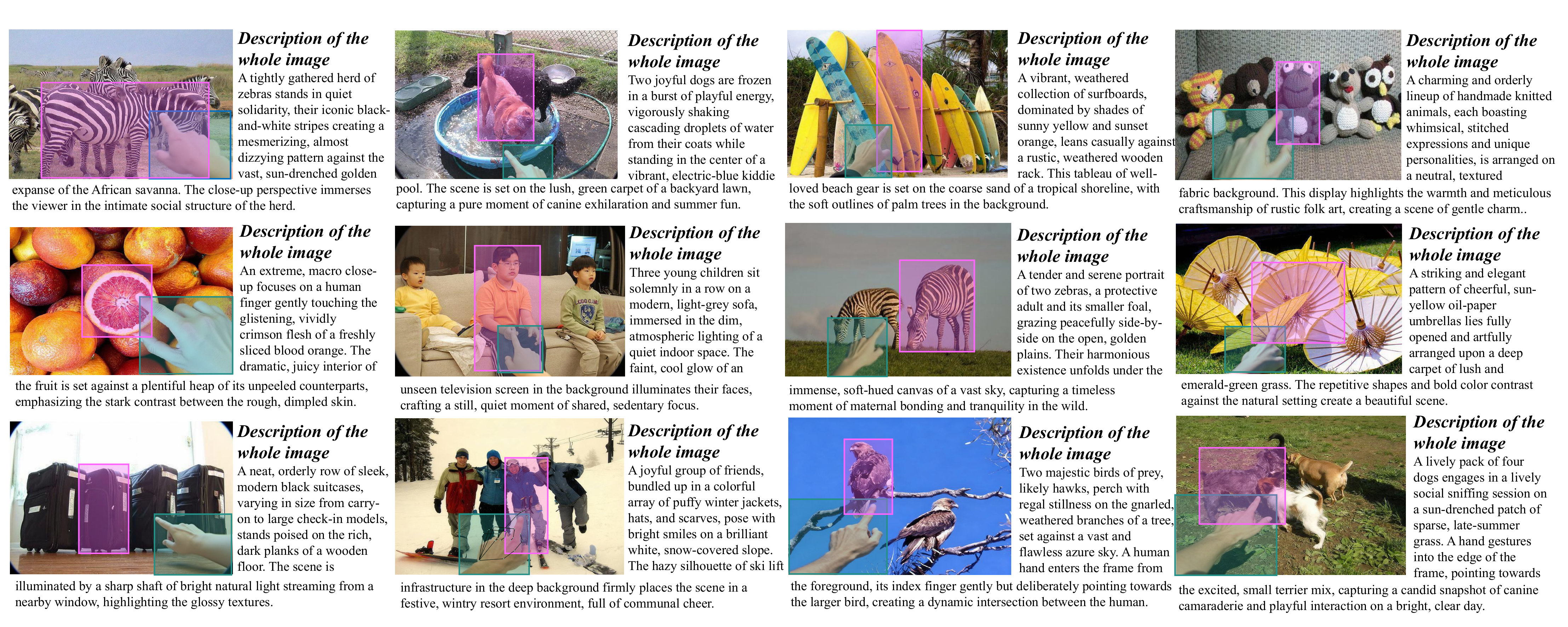}
    \caption{Display of Randomly Sampled Image Captions.}
    \label{fig:caption supple}
\end{figure*}

\section{Dataset Statistics and Composition}

Our dataset is designed to cover complex human-computer interaction scenarios from a First-Person View. The dataset comprises several dozen independent scenarios, encompassing hundreds of different object and action categories.

The data sources are divided into three parts, as shown in Table \ref{tab:ratio}.

\begin{table}[h]
    \centering
    \caption{Data Source Composition}
    \label{tab:data_source}
    \begin{tabular}{c|c}
        \hline
        \textbf{Source} & \textbf{Ratio} \\
        \hline
        Real-World Data & {39.35\%} \\
        Synthetic Generation & {38.96\%} \\
        Image Editing & {21.69\%} \\
        \hline
    \end{tabular}
    \label{tab:ratio}
\end{table}

The dataset mainly includes: home scenarios, campus scenarios, daily interaction scenarios, supermarket/mall interaction scenarios, and others. For the detailed categories and corresponding IDs of the dataset, please refer to Appendix Table \ref{tab:object_categories_six_column}.

To provide a comprehensive view of the dataset's features, we have augmented the data presentation with detailed examples. Figure \ref{fig:dataset show supple} offers a representative showcase of the dataset's augmented data. Specifically, Figure \ref{fig:QA supple} illustrates Visual Question Answering (VQA) examples for the pointed target objects, while Figure \ref{fig:caption supple} presents image captioning examples for the entire scene.

\subsection{Data Annotation Methodology}

The dataset annotation adopts a hybrid model combining manual fine-grained annotation with algorithmic pre-annotation. The annotators are mainly researchers from various universities who possess rich expertise in computer vision. Among the annotators, males account for 75\%, and females account for 25\%.

\subsection{Data Acquisition Details}

The manual collection part of the dataset primarily utilizes RayNeo smart glasses (Model: RayNeo V3).
The key parameters of this device are as shown in Table \ref{tab:RayNeo V3}.

\begin{table*}[h]
    \centering
    \caption{Key Parameters of RayNeo V3}
    \label{tab:rayneo_v3_params}
    \begin{tabular}{c|c}
        \hline
        \textbf{Parameter} & \textbf{Specification} \\
        \hline
        Model & RayNeo V3 \\
        Sensor & Sony IMX681, 12MP \\
        Lens & 5P customized F2.2 large aperture 16mm ultra-wide-angle \\
        Algorithm & ArcSoft customized algorithm \\
        ISP & 1st generation Snapdragon AR1 dual ISP \\
        \hline
    \end{tabular}
    \label{tab:RayNeo V3}
\end{table*}
\begin{table*}[htp]
\centering
\caption{Model Specifications and Evaluation Metrics. (Test on Egocentric Deictic Visual Grounding - EDG)}
\label{tab:models_detail}
\resizebox{0.95\linewidth}{!}{%
\begin{tabular}{c|c|c|c|c}
\hline
\textbf{Model Name} & \textbf{Model Size} & \textbf{Inference Time (h:min)} & \textbf{Peak VRAM Usage} & \textbf{Model Source} \\
\hline
Qwen3-VL-8B                          & 8.78 B & 1:47 & 16.6 GB & Open-Sourced Model \\
Florence-2-large                     & 0.77 B & 0:36 & 1.8 GB & Open-Sourced Model \\
Ferret                               & 13.37 B & 4:52 & 26.9 GB & Open-Sourced Model \\
LLaVA-OneVision-1.5-8B-Instruct      & 8.53 B & 1:58 & 16.3 GB & Open-Sourced Model \\
GroundingGPT                         & 7.90 B & 1:52 & 16.5 GB & Open-Sourced Model \\
InternVL-3.5-8B                      & 7.91 B & 2:21 & 40.2 GB & Open-Sourced Model \\
\hline
\end{tabular}%
}
\vspace{-5pt}
\end{table*}
All collected data has undergone privacy protection processing. Areas involving facial or Personally Identifiable Information (PII) have been blurred to ensure compliance with data ethics requirements.

\section{Experiment Details}
\label{sec:experiment_details}

We have organized the code for our method in the \texttt{codes} directory. Upon publication, we will release the code and pretrained models. Detailed usage instructions are provided in \texttt{/codes/EgoPoint-Ground\_code/README.md}.

\subsection{Experimental Settings} 
 All experiments are conducted with the official task scripts in \texttt{code\_for\_evaluate\_on\_Ego} under inference-only settings (no task-specific fine-tuning). Across REC/POG/EDG/D-VQA, each script loads model checkpoints via \texttt{from\_pretrained(..., device\_map="auto")} and runs mixed-precision inference (typically \texttt{bf16} in the provided \texttt{run.sh} files). The arguments \texttt{--start/--end} are used to control the evaluated sample range (\texttt{end=-1} indicates the full split). 
 
For REC and EDG, the target instance is sampled from the non-hand annotations; each sample is evaluated over its \texttt{underspecified\_referent} entries (typically three expressions), and each expression is treated as an independent test case. For POG and EDG, the hand box is taken from \texttt{ann\_id=anno\_hand}, and the GT pointed object is selected from non-hand annotations. For D-VQA, the hand box is also taken from \texttt{ann\_id=anno\_hand}, and the GT category is extracted from \texttt{category\_name} in non-hand annotations. 
 
 Prompt templates are task-specific but parser-aligned: REC uses underspecified text only, POG uses hand bbox only, EDG uses both hand bbox and underspecified text, and D-VQA uses hand bbox with a question (e.g., "what is this?") to prompt category prediction of the pointed object. Coordinate formats are adapted per model/script (\texttt{absolute}, \texttt{relative\_1}, or \texttt{relative\_1000}), then converted to a unified absolute box space for scoring. Most scripts use deterministic decoding (\texttt{do\_sample=False}); model-specific official settings are preserved where required (e.g., sampling-based decoding in some GroundingGPT/DeepSeek-VL2 scripts). All scripts enforce short final bbox outputs (instead of long reasoning traces) to ensure stable parsing and fair comparison. 
 
\subsection{Other Details in Experiments}
To support reproducibility, each evaluation JSON includes both run metadata and per-sample records. Metadata contains model path, data path, coordinate mode, sample count, parsed prediction count (\texttt{parsed\_ok}), valid IoU count, and skip/error statistics (e.g., \texttt{missing\_target}, \texttt{missing\_hand\_bbox}, \texttt{parse\_failed}, \texttt{infer\_exception}). Per-sample entries store GT/predicted boxes in both absolute and relative formats (when available), raw model outputs, and error tags.

Box extraction is implemented with a JSON-first parser (e.g., \texttt{\{"bbox\_2d": [x1, y1, x2, y2]\}}) plus regex fallback for list-form outputs. Predicted and GT boxes are sanitized, clamped to image boundaries, and converted across coordinate systems before metric computation. Metrics are uniformly reported as \texttt{IoU@0.3}, \texttt{IoU@0.5}, \texttt{IoU@0.7}, and \texttt{IoU\_avg}. In addition, scripts save qualitative diagnostics in \texttt{visualize/<model\_name>/}, including overlay images (GT/pred/hand when applicable) and paired prompt/raw-output text files, which support error analysis and result verification.

To comprehensively evaluate the performance of general-purpose models on our dataset, we summarize key efficiency metrics for each baseline, including inference latency, peak VRAM consumption, and model parameter count, as presented in Table \ref{tab:models_detail}. Furthermore, to ensure experimental fairness and reproducibility, we detail the specific input paradigms employed for each model across the D-REC and POG tasks in Tables \ref{tab:rec_input} and Table \ref{tab:pog input}, respectively.
\begin{table*}[htp]
\centering
\caption{Model Input Paradigms (Test on Language-Only Grounding (D-REC))}
\label{tab:rec_input} 
\renewcommand{\arraystretch}{1.3} 
\begin{tabularx}{\textwidth}{@{} l @{\hspace{2em}} X @{}}
\toprule
\textbf{Model Name} & \textbf{Input Paradigm} \\
\midrule
GroundingGPT & 
\texttt{"[object caption]"} \\
\midrule
Qwen3-VL & 
\texttt{"You are given an egocentric (first-person) image where a visible hand/finger is pointing to an object. Your task is to localize the pointed-at target object and output its bounding box. Use \texttt{[x1, y1, x2, y2]} coordinates in range \texttt{[0, 1000]} (relative to image size). The underspecified reference below is auxiliary text about the target. Underspecified reference: \texttt{\{underspecified\_referent\}}. Return ONLY one JSON object in this format: \texttt{\{"bbox\_2d": [x1, y1, x2, y2]\}}."} \\
\midrule
InternVL-3.5 & 
\texttt{"You are given an egocentric (first-person) image where a visible hand/finger is pointing to an object, predict the bounding box of \texttt{<ref>\{underspecified\_referent\}</ref>} being pointed at in format \texttt{[x1, y1, x2, y2]} where coordinates are in range \texttt{[0, 1000]} (relative to image size). Return ONLY the bbox of the pointed object in format \texttt{[x1, y1, x2, y2]}. Give me the result directly, do not reason."} \\
\midrule
LLaVA-OneVision & 
\texttt{"You are given an egocentric (first-person) image where a visible hand/finger is pointing to an object. Your task is to localize the pointed-at target object and output its bounding box, where each coordinate is normalized to \texttt{[0, 1]} relative to image width/height. Underspecified reference: \texttt{\{underspecified\_referent\}}. Return ONLY the bbox in this exact format: \texttt{[x1, y1, x2, y2]}."} \\
\midrule
Ferret & 
\texttt{"You are given an egocentric (first-person) image where a visible hand/finger is pointing to an object. Your task is to localize the pointed-at target object and output its bounding box in format \texttt{[x1, y1, x2, y2]}, where coordinates are in range \texttt{[0, 1000]} (relative to image size). Underspecified reference: \texttt{\{underspecified\_referent\}}. Return ONLY the bbox of the pointed object as \texttt{[x1, y1, x2, y2]}."} \\
\midrule
DeepSeek-VL2 & 
\texttt{"<|grounding|> You are given an egocentric (first-person) image where a visible hand/finger is pointing to an object. Your task is to localize the pointed-at target object and output its bounding box in format \texttt{[x1, y1, x2, y2]}, where each coordinate is in \texttt{[0, 1000]} relative to image size. Target object's underspecified reference: \texttt{\{underspecified\_referent\}}. Return ONLY one bbox in \texttt{[x1, y1, x2, y2]}."} \\
\bottomrule
\end{tabularx}
\label{tab:rec input}
\vspace{-5pt}
\end{table*}

\begin{table*}[htp]
\centering
\caption{Model Input Paradigms (Test on Pointing Only Grounding - POG)}
\label{tab:pog_input} 
\renewcommand{\arraystretch}{1.3} 
\begin{tabularx}{\textwidth}{@{} l @{\hspace{2em}} X @{}}
\toprule
\textbf{Model Name} & \textbf{Input Paradigm} \\
\midrule
GroundingGPT & 
\texttt{"You are a pointed object bounding box detector. Given an image and a hand bbox, predict the bounding box of the object being pointed at in format \texttt{[x1, y1, x2, y2]}. Hand bbox (xyxy, normalized to \texttt{[0,1]}): \texttt{\{hand\_input\}}. Return ONLY the bbox of the pointed object in this format: \texttt{[x1, y1, x2, y2]}."} \\
\midrule
Qwen3-VL & 
\texttt{"You are a pointed object bounding box detector. Given an image and a hand bbox, predict the bounding box of the object being pointed at in format \texttt{[x1, y1, x2, y2]} where coordinates are in range \texttt{[0, 1000]} (relative to image size). Input hand bbox: \texttt{\{hand\_input\}}. Return ONLY the bbox of the hand-pointed object in this format: \texttt{\{"bbox\_2d": [x1, y1, x2, y2]\}}."} \\
\midrule
InternVL-3.5 & 
\texttt{"You are a pointed object bounding box detector. Given an image and a hand bbox, predict the bounding box of \texttt{<ref>the object being pointed</ref>} in format \texttt{[x1, y1, x2, y2]} where coordinates are in range \texttt{[0, 1000]} (relative to image size). Input hand bbox: \texttt{\{hand\_input\}}. Return ONLY the bbox of the pointed object in format \texttt{[x1, y1, x2, y2]}."} \\
\midrule
LLaVA-OneVision & 
\texttt{"You are a pointed object bounding box detector. Given an image and a hand bbox, predict the bounding box of the object being pointed at in format \texttt{[x1, y1, x2, y2]}, where each coordinate is normalized to \texttt{[0, 1]} relative to image width/height. Input hand bbox (xyxy, normalized to \texttt{[0, 1]}): \texttt{\{hand\_input\}}. Return ONLY the bbox in this exact JSON format: \texttt{[x1, y1, x2, y2]}."} \\
\midrule
Ferret & 
\texttt{"You are a pointed object bounding box detector. Given an image and a hand bbox, predict the bounding box of the object being pointed at in format \texttt{[x1, y1, x2, y2]} where coordinates are in range \texttt{[0, 1000]} (relative to image size). Input hand bbox: \texttt{\{hand\_input\}}. Return ONLY about the POINTED OBJECT, do not mention the hand, do not reason, return the result directly as \texttt{[x1, y1, x2, y2]}."} \\
\midrule
DeepSeek-VL2 & 
\texttt{"<|grounding|> You are a pointed object grounding model. Given an image and a hand bbox, predict the bbox of the object being pointed at in format \texttt{[x1, y1, x2, y2]} where coordinates are in \texttt{[0, 1000]} relative to image size. Input hand bbox: \texttt{\{hand\_input\}}. Return ONLY one bbox in \texttt{[x1, y1, x2, y2]}."} \\
\bottomrule
\end{tabularx}
\label{tab:pog input}
\vspace{-5pt}
\end{table*}

\begin{table*}[h!]
    \centering
    \caption{Detailed Mapping of Dataset Object Categories to Their Corresponding Integer IDs.}
    \label{tab:object_categories_six_column}
    \resizebox{1.05\textwidth}{!}{
    \begin{tabular}{lc|lc|lc|lc|lc|lc}
        \hline
        \textbf{Category} & \textbf{ID} & \textbf{Category} & \textbf{ID} & \textbf{Category} & \textbf{ID} & \textbf{Category} & \textbf{ID} & \textbf{Category} & \textbf{ID} & \textbf{Category} & \textbf{ID} \\
        \hline
        advertising board & 0 & airplane & 1 & alarm & 2 & ammeter & 3 & apple & 4 & artwork & 5 \\
        ashtray & 6 & audio equipment & 7 & axe & 8 & backpack & 9 & bag & 10 & ball & 11 \\
        ball of thread & 12 & ballpoint pen & 13 & banana & 14 & band-aid & 15 & baseball bat & 16 & basketball & 17 \\
        bear & 18 & bed & 19 & beer & 20 & bell & 21 & belt & 22 & bench & 23 \\
        beverage & 24 & bicycle & 25 & bike & 26 & bird & 27 & blackboard & 29 & blanket & 30 \\
        board & 31 & boat & 32 & book & 33 & bookshelf & 34 & bottle & 35 & bouquet of flowers & 36 \\
        bowl & 37 & box & 38 & bread & 39 & broccoli & 40 & brush & 41 & bucket & 42 \\
        bus & 43 & bus stop & 44 & button & 45 & cabinet & 46 & cake & 47 & calculator & 48 \\
        calendar & 49 & camera & 50 & can & 51 & candy & 52 & car & 53 & card & 54 \\
        carrot & 55 & castle & 56 & cat & 57 & cattle & 58 & cd & 59 & cell phone & 60 \\
        chain & 61 & chair & 62 & chandelier & 63 & charger & 64 & chicken & 65 & chocolate & 66 \\
        church & 67 & circuit board & 68 & cling film & 69 & clock & 70 & cloth & 71 & clothes & 72 \\
        coat & 73 & coffee & 74 & coffee machine & 75 & coin & 76 & corn & 77 & cosmetics & 78 \\
        cotton swab & 79 & couch & 80 & cow & 81 & crane & 82 & cup & 83 & curtain & 84 \\
        dining table & 85 & disc & 87 & dog & 88 & doll & 89 & donut & 90 & door & 91 \\
        door handle & 92 & duck & 93 & dumbbell & 94 & ear stud & 95 & egg & 96 & electric drill & 97 \\
        elephant & 98 & elevator & 99 & eye drops & 100 & fan & 101 & fence & 102 & file & 103 \\
        fire engine & 104 & fire extinguisher & 105 & fire hydrant & 106 & fireplace & 107 & fireworks & 108 & fish & 109 \\
        fishhook & 110 & fishing hook & 111 & fishing line & 112 & fishing rod & 113 & fitness equipment & 114 & flag & 115 \\
        flashlight & 116 & flour & 117 & flower & 118 & flowerpot & 119 & folding chair & 120 & food & 121 \\
        fork & 122 & fragrance device & 123 & fries & 124 & frisbee & 125 & galvanometer & 126 & game boy & 127 \\
        game console & 128 & game controller & 129 & game disc & 130 & giraffe & 131 & glasses & 132 & globe & 133 \\
        gloves & 134 & goat & 135 & goblet & 136 & golf club & 137 & grape & 138 & green onion & 139 \\
        guitar & 140 & hair drier & 141 & hair dryer & 142 & hamburger & 143 & hand & 144 & handbag & 145 \\
        hat & 146 & head & 147 & headphones & 148 & helmet & 149 & highball glass & 150 & hole & 151 \\
        hook & 152 & horse & 153 & hot dog & 154 & hula hoop & 155 & iron hammer & 156 & jam & 157 \\
        jeans & 158 & jewelry & 159 & juice & 160 & juicer & 161 & kettle & 162 & key & 163 \\
        keyboard & 164 & kid & 165 & kite & 166 & knee pads & 167 & knife & 168 & lamp & 169 \\
        laptop & 170 & leaf & 171 & led lamp & 172 & lemon & 173 & lens cap & 174 & light bulb & 175 \\
        lighthouse & 176 & lipstick & 177 & lock & 178 & lubricating oil & 179 & machine & 180 & magazine & 181 \\
        mailbox & 182 & meat & 183 & medicine & 184 & microwave & 185 & milk & 186 & mirror & 187 \\
        model & 188 & monitor & 189 & monument & 190 & motorcycle & 191 & mouse & 192 & nail & 193 \\
        nail art sheet & 194 & necklace & 195 & needle & 196 & net & 197 & newspaper & 198 & notebook & 199 \\
        nut & 200 & orange & 202 & oven & 203 & overcoat & 204 & painting & 205 & pan & 206 \\
        paper & 207 & paper bag & 208 & parking meter & 209 & pen & 211 & pen holder & 212 & pencil & 213 \\
        pepper powder & 215 & person & 216 & phone & 217 & phone booth & 218 & photo & 219 & piano & 220 \\
        pigment & 221 & pigment tray & 222 & pillar & 223 & pillow & 224 & pizza & 225 & pizza box & 226 \\
        plant & 227 & plate & 228 & pliers & 229 & porcelain & 230 & pot & 231 & potted plant & 232 \\
        rack & 233 & radio & 234 & reagent bottle & 235 & red-crowned crane & 236 & refrigerator & 237 & remote & 238 \\
        rice & 239 & rice cooker & 240 & road sign & 241 & rockery & 242 & rose & 243 & sachs & 244 \\
        salt & 245 & sandals & 246 & sandwich & 247 & sausage & 248 & saw & 249 & scale & 250 \\
        scarf & 251 & scissors & 252 & screw & 253 & screwdriver & 254 & sculpture & 255 & seasoning & 256 \\
        shampoo & 257 & sheep & 258 & shoes & 259 & shop & 260 & shopping cart & 261 & showerhead & 262 \\
        signboard & 263 & signpost & 264 & sink & 265 & skateboard & 266 & skis & 267 & slippers & 268 \\
        snack & 269 & snowboard & 270 & soap & 271 & socket & 272 & sofa & 273 & soup & 274 \\
        soy sauce & 275 & spoon & 276 & sports ball & 277 & squirrel & 278 & stapler & 279 & star & 280 \\
        stick & 281 & sticker & 282 & sticky note & 283 & stirring rod & 284 & stone & 285 & strawberry & 287 \\
        string & 288 & suitcase & 289 & sunflower & 290 & sunglasses & 291 & surfboard & 292 & t-shirt & 293 \\
        table & 294 & tape measure & 295 & teddy bear & 296 & tennis racket & 297 & test tube & 298 & tie & 299 \\
        timer & 300 & tire & 301 & toaster & 302 & toilet & 303 & tomato & 304 & tools & 305 \\
        toothbrush & 306 & toothpaste & 307 & towel rack & 308 & toy & 309 & toy house & 310 & trachea & 311 \\
        track & 312 & traffic light & 313 & train & 314 & transformer & 315 & tree & 316 & trousers & 317 \\
        truck & 318 & trunk & 319 & tv & 320 & umbrella & 321 & usb flash drive & 322 & vase & 323 \\
        vending machine & 325 & violin & 326 & vitamin & 327 & vr glasses & 328 & wall & 329 & wallet & 330 \\
        water & 331 & watering pot & 332 & watermelon & 333 & whiteboard & 334 & window & 335 & wine & 336 \\
        wine bottle & 337 & wine glass & 338 & wire & 339 & wooden block & 340 & wooden board & 341 & wool ball & 342 \\
        wrench & 343 & yoga wheel & 344 & zebra & 345 & hot water bottle & 346 & ruler & 347 & chopsticks & 348 \\
        chopstick & 349 & conditioner & 350 & & & & & & & & \\
        \hline
    \end{tabular}
    }
\end{table*}
\end{document}